\documentclass[letterpaper,numbers,compress]{article} 
    \PassOptionsToPackage{numbers,compress}{natbib}
    \usepackage[preprint]{neurips_2024}

\usepackage[utf8]{inputenc} %
\usepackage[T1]{fontenc}    %
\usepackage{hyperref}       %
\usepackage{url}            %
\usepackage{booktabs}       %
\usepackage{amsfonts}       %
\usepackage{nicefrac}       %
\usepackage{microtype}      %
\usepackage{xcolor}         %

\usepackage{graphicx}

\usepackage{bm}
\usepackage{amsmath, amsthm, amssymb}

\usepackage{centernot}%
\newcommand{\adjacent}[1][.7em]{\mathrel{\rule[.5ex]{#1}{.4pt}}}
\newcommand{\notadjacent}[1][.7em]{\mathrel{\centernot{\adjacent[#1]}}}
\title{An efficient search-and-score algorithm for ancestral graphs using multivariate information scores}

\author{Nikita~Lagrange,  ~Herv\'e Isambert\\
  CNRS, UMR168, Institut Curie, PSL University, Sorbonne University, Paris, France\\
  \texttt{nikita.lagrange@curie.fr, herve.isambert@curie.fr}}

\begin{document}

\maketitle

\begin{abstract}
  We propose a greedy search-and-score algorithm for ancestral graphs, which include directed as well as bidirected edges, originating from unobserved latent variables. The normalized likelihood score of ancestral graphs is estimated in terms of multivariate information over relevant ``$ac$-connected subsets'' of vertices, $\bm{C}$, that are connected through collider paths confined to the ancestor set of $\bm{C}$. For computational efficiency, the proposed two-step algorithm relies on local information scores limited to the close surrounding vertices of each node (step 1)  and edge (step 2). This computational strategy, although restricted to information contributions from $ac$-connected subsets containing up to two-collider paths, is shown to outperform state-of-the-art causal discovery methods on challenging benchmark datasets.
\end{abstract}

\section{Introduction}

\noindent
The likelihood function plays a central role in the selection of a graphical model  $\mathcal{G}$ based on observational data $\mathcal{D}$.
Given $N$ independent samples from $\mathcal{D}$, the likelihood   ${\mathcal{L_{D\vert G}}}$ that they might have been generated by the graphical model $\mathcal{G}$ is given by~\cite[]{koller}, %
\begin{eqnarray}
{\mathcal{L_{D\vert G}}}={1\over Z_{\mathcal{D,G}}}\;{\exp\Big({-N H({p,q})}\Big)}\;\;\;\;%
\label{likelihood}
\end{eqnarray}
where $H({p,q})=-\sum_{\bm{x}}p(\bm{x})\log q(\bm{x})$ is the cross-entropy between the empirical probability distribution $p(\bm{x})$ of the observed data $\mathcal{D}$ and the theoretical probability distribution $q(\bm{x})$ of the model $\mathcal{G}$  and $Z_{\mathcal{D,G}}$ a data- and model-dependent factor ensuring proper normalization condition for finite dataset.
In short, 
Eq.\ref{likelihood} 
results from the asymptotic probability that the $N$ independent samples,  $\bm{x}^{(1)},\cdots,\bm{x}^{(N)}$, are drawn from the model distribution, $q(\bm{x})$, {\em i.e.}~\mbox{${\mathcal{L_{D\vert G}}}\equiv q(\bm{x}^{(1)},\cdots,\bm{x}^{(N)})=\prod_iq(\bm{x}^{(i)})$}, rather than the empirical distribution, $p(\bm{x})$.
This leads to, \mbox{$\log {\mathcal{L_{D\vert G}}}=\sum_i\!\log q(\bm{x}^{(i)})$}, which converges towards $N\sum_{\bm{x}} p(\bm{x})\log q(\bm{x})= -N\,H({p,q})$ in the large sample size limit, $N\to \infty$, with $\log Z_{\mathcal{D,G}} = {\mathcal{O}}(\log N)$.

The structural constraints of the model $\mathcal{G}$ 
translate into
the factorization form of the theoretical probability distribution, $q(\bm{x})$ \cite[]{pearl1985,pearl_book1988,pearl_book2009,spirtes_book2000,richardson2009}. 
In particular,  the probability distribution of Bayesian networks (BN) factorizes in terms of conditional probabilities of each variable given its parents, as $q_{_{\rm\tiny BN}}(\bm{x})=\prod_{i}q(x_i\vert \mathbf{pa}_{X_i})$, where $\mathbf{pa}_{X_i}$ denote the values of the parents of node $X_i$ in $\mathcal{G}$,  $\mathbf{Pa}_{X_i}$.
For  Bayesian networks, the factors of the model distribution, $q(x_i\vert \mathbf{pa}_{X_i})$, can be directly estimated with the empirical conditional probabilities of each node given its parents as, $q(x_i\vert \mathbf{pa}_{X_i})\equiv p(x_i\vert \mathbf{pa}_{X_i})$, leading to the well known estimation of the likelihood function in terms of conditional entropies $H(X_i\vert \mathbf{Pa}_{X_i})=-\sum_{\bm{x}}p(x_i,\mathbf{pa}_{X_i})\log p(x_i\vert \mathbf{pa}_{X_i})$,
\begin{equation}
{\mathcal{L}_{\mathcal{D\vert G}_{\rm\tiny BN}}}\!={1\over Z_{\mathcal{D,G}_{_{\rm\tiny BN}}}}\;{\exp{\Big(\!-N\!\!\!\sum_{X_i\in\bm{V}}^{\rm vertices}\!\!\!H(X_i\vert \mathbf{Pa}_{X_i})\Big)}}\label{BN}
\end{equation}
This paper concerns the experimental setting for which some variables of the underlying Bayesian model are not observed. This frequently occurs in practice for many applications. 
We derive an explicit likelihood function for the
class of ancestral graphs,
which include directed as well as bidirected edges, 
arising from the presence of unobserved latent variables. 
Tian and Pearl 2002 \cite[]{tian2002} showed that the
probability distribution of such graphs
factorizes into c-components including subsets of variables connected through bidirected paths ({\it i.e.}~containing only bidirected edges).
Richardson 2009 \cite[]{richardson2009} later proposed a refined factorization of the model distribution of the broader class of acyclic directed mixed graphs in terms of conditional probabilities over ``head'' and ``tail'' subsets of variables within each ancestrally closed subsets of vertices.
However, unlike with Bayesian networks, the contributions of c-components or head-and-tail factors to the likelihood function cannot simply be estimated in terms of empirical distribution $p(\bm{x})$, as shown below.
This leaves the likelihood function of ancestral graphs difficult to estimate from empirical data, in general, although iterative methods have been developped when the data is %
normally distributed \cite[]{richardson2002,drton2009b,evans2010,triantafillou2016,rantanen2021,claassen2022}.

The present paper
{\color{black}{provides an explicit decomposition of the likelihood function of
ancestral graphs
in terms of multivariate cross-information over relevant `$ac$-connected' subsets of variables, Figs.~\ref{fig:CrossEntropyDecomposition_all}.}, which
do not rely 
  on the head-and-tail factorization but coincide with the parametrizing sets \cite[]{hu2020} derived from the head-and-tail factorization. It suggests a natural estimation of these revelant contributions to the likelihood function in terms of empirical distribution $p(\bm{x})$.}
This result extends the likelihood expression of Bayesian Networks (Eq.~\ref{BN}) to include the effect of unobserved latent variables %
and enables the implementation of a greedy search-and-score algorithm for ancestral graphs.
For computational efficiency, the proposed two-step algorithm relies on local information scores limited to the close surrounding vertices of each node (step 1) and edge (step 2). This computational strategy is shown to outperform state-of-the-art causal discovery methods on challenging benchmark datasets.

\section{Theoretical results}

\vspace*{-0.1cm}

\subsection{Multivariate cross-entropy and cross-information}
The theoretical result of the paper (Theorem~1) is expressed in terms of multivariate cross-information derived from multivariate cross-entropies through the Inclusion-Exclusion Principle. 
The same expressions can be written between multivariate information and multivariate entropies by simply substituting $q(\{x_i\})$ with $p(\{x_i\})$ in the equations below and will be used to estimate the likelihood function of ancestral graphs (Proposition~3).

As recalled above, the cross-entropy  between $m$ variables, $\bm{V}=\{X_1,\cdots,X_m\}$, is defined as,\vspace*{-0.5cm} %

\begin{eqnarray}
H(\bm{V})\!\!\!&=&\!\!\!-\!\sum_{\{x_i\}} p(x_1,\cdots,x_m) \log{{ q(x_1,\cdots,x_m)\;\;\;\;\;\;}}
\label{GE}
\end{eqnarray}
\vspace*{-0.4cm}

\noindent 
where $p(\{x_i\})$ is the empirical joint probability distribution of the variables $\{X_i\}$ and $q(\{x_i\})$ the joint probability distribution of the model.
Bayes formula, $q(\{x_i\},\{y_j\})=q(\{x_i\}|\{y_j\})\;q(\{y_j\})$, directly translates into the definition of conditional cross-entropy through the decomposition,
\begin{equation}
H(\{X_i\},\{Y_j\})=H(\{X_i\}|\{Y_j\})+H(\{Y_j\})
\label{GEbayes}
\end{equation}
Multivariate (cross) information, $I(\bm{V})\equiv I(X_{1};\cdots;X_{m})$, 
are defined from multivariate (cross) entropies %
through Inclusion-Exclusion formulas %
over all subsets of variables 
\cite[]{mcgill1954,ting1962,han1980,yeung1991}
as, %
\begin{eqnarray}
I(X)&=& H(X)\nonumber\\
I(X;Y)&=& H(X)+H(Y)-H(X,Y)\nonumber\\
I(X;Y;Z)&=& H(X)+H(Y)+H(Z)-\!H(X,Y) -\!H(X,Z)-\!H(Y,Z)+\!H(X,Y,Z)\nonumber\\
I(\bm{V})&=&-\sum_{\bm{S}\subseteq \bm{V}}(-1)^{|\bm{S}|}H(\bm{S}) %
\label{GIE0}
\end{eqnarray}
where the semicolon separators are needed to distinguish multipoint (cross) information from joint variables as $\{X,Z\}$ in $I(\{X,Z\};Y)=I(X;Y)+I(Z;Y) -I(X;Y;Z)$.
 Below, implicit separators between non-conditioning variables in multivariate (cross) information will always correspond to semicolons, {\em e.g.}~as in $I(\bm{V})$ in Eq.~\ref{GIE0}.  
Unlike multivariate (cross) entropies, which are always positive, $H(X_1,\cdots,X_k)\geqslant 0$, multivariate (cross) information, $I(X_1;\cdots;X_k)$, can be positive or negative for \mbox{$k\geqslant 3$}, while they remain always positive for $k<3$, {\em i.e.}~$I(X;Y)\geqslant 0$ and $I(X)\geqslant 0$.

In turn, multivariate (cross) entropies can be expressed through the Principle of Inclusion-Exclusion into the same expression form but in terms of multivariate (cross) information, %
\begin{eqnarray}
H(\bm{V})&=&-\sum_{\bm{S}\subseteq \bm{V}}(-1)^{|\bm{S}|}I(\bm{S}),
\label{MIE}
\end{eqnarray}
Conditional multivariate (cross) information $I(\bm{V}|Z)$ 
are defined similarly as multivariate (cross) information $I(\bm{V})$ but in terms of conditional (cross) entropies as,
\begin{eqnarray}
I(\bm{V}|Z)&=&-\sum_{\bm{S}\subseteq \bm{V}}(-1)^{|\bm{S}|}H(\bm{S}|Z) %
\label{GIE}
\end{eqnarray}
Eqs.~\ref{GIE0}~\&~\ref{GIE} lead to a decomposition rule relative to a variable $Z$, Eq.~\ref{DecompositionRule1}, which can be conditioned on a set of joint variables, \mbox{$\bm{A}=\{A_1,\cdots,A_m\}$}, with implicit comma separators for conditioning variables in Eq.~\ref{DecompositionRule2},
\begin{eqnarray}
I(\bm{V})&\!\!\!=\!\!\!& I(\bm{V}|Z)+I(\bm{V};Z)\label{DecompositionRule1}\\
I(\bm{V}|\bm{A})&\!\!\!=\!\!\!& I(\bm{V}|Z,\bm{A})+I(\bm{V};Z|\bm{A})\;\;\;\;
\label{DecompositionRule2}
\end{eqnarray}
Alternatively, conditional (cross) information, such as $I(X;Y|\bm{A})$, can be expressed in terms of non-conditional (cross) entropies using Eq.~\ref{GEbayes},
\begin{eqnarray}
I(X;Y|\bm{A})&\!\!\!=\!\!\!&H(X|\bm{A})+H(Y|\bm{A})-H(X,Y|\bm{A}) \nonumber\\
&\!\!\!=\!\!\!&H(X,\bm{A})+H(Y,\bm{A})-H(X,Y,\bm{A})-H(\bm{A}) %
\label{CMIE0}
\end{eqnarray}
which can in turn be expressed in terms of  non-conditional  (cross) information as,
\begin{eqnarray}
I(X;Y|\bm{A})\!\!\!\!&=&\!\!\!\!I(X;Y)-\cdots(-1)^{k}\!\!\!\!\!\sum_{{i_1}< \cdots <  {i_k}} \!\!\!\!I(X;Y;A_{i_1};\cdots\!;A_{i_k})+\cdots (-1)^{m} I(X;Y;A_{1};\cdots\!;A_{m})  \nonumber\\
\!\!\!&=&\!\!\!\!\sum_{\bm{S^\prime}\subseteq \bm{S}}^{X,Y\in \bm{S^\prime}}(-1)^{|\bm{S^\prime}|}I(\bm{S^\prime}),
\label{CMIE}
\end{eqnarray}
where $\bm{S}=\{X,Y\}\cup\bm{A}$. This corresponds, up to an opposite sign, to {\em all (cross) information terms including both $X$ and $Y$} in the expression of the multivariate  (cross) entropy, $H(X,Y,\bm{A})$, Eq.~\ref{MIE}. 

\vspace*{0.2cm}

\subsection{Graphs and connection criteria} 
\subsubsection{Directed mixed graphs and ancestral graphs}

Two vertices are said to be {\bf adjacent} if there is an edge (of any type) between them, $X *\!\!\adjacent\!\!* Y$, where $*$ stands for any (head or tail) end mark. $X$ and $Y$ are said to be {\bf neighbors} if $X\adjacent Y$, {\bf parent} and {\bf child} if  $X\rightarrow Y$ and {\bf spouses} if $X\longleftrightarrow Y$ in $\mathcal{G}$.

A {\bf path} in  $\mathcal{G}$ is a sequence of distinct vertices $V_1,\ldots,V_n$ consecutively adjacent in $\mathcal{G}$, as, \mbox{$V_1 *\!\!\adjacent\!\!* V_2 *\!\!\adjacent\!\!* \cdots *\!\!\adjacent\!\!* V_{n-1} *\!\!\adjacent\!\!* V_n$}. In particular, a {\bf collider path} between $V_1$ and $V_n$  has the form \mbox{$V_1 *\!\!\rightarrow V_2 \longleftrightarrow \cdots  \longleftrightarrow V_{n-1} \leftarrow\!\!* V_n$} and a {\bf directed path}  corresponds to \mbox{$V_1 \rightarrow V_2 \rightarrow \cdots \rightarrow  V_n$}.

$X$ is called an {\bf ancestor} of $Y$ and $Y$ a {\bf descendant} of $X$ if $X=Y$ or there is a {\bf directed path} from $X$ to $Y$, $X\rightarrow \cdots \rightarrow  Y$.  $\mathbf{An}_\mathcal{G}(Y)$ denotes the {\bf set of ancestors} of $Y$ in $\mathcal{G}$. By extension, for any subset of vertices, $\bm{C}\subseteq \bm{V}$, $\mathbf{An}_\mathcal{G}(\bm{C})$ denotes the set of ancestors for all $Y\!\in\!\bm{C}$ in $\mathcal{G}$.

A {\bf directed mixed graph} is a vertex-edge graph $\mathcal{G}=(\bm{V},\bm{E})$ that can contain two types of edges: %
directed ($\rightarrow$) and bidirected ($\longleftrightarrow$) edges.

A {\bf directed cycle} occurs in $\mathcal{G}$ when $X\in \mathbf{An}_\mathcal{G}(Y)$ and $X\leftarrow Y$. An {\bf almost directed cycle} occurs when $X\in \mathbf{An}_\mathcal{G}(Y)$ and $X\longleftrightarrow Y$.

\noindent
{\bf Definition 1.} An {\bf ancestral graph} is a directed mixed graph:\\
\hspace*{1cm}{\em i)} ~~without directed cycles;\\
\hspace*{1cm}{\em ii)} ~without almost directed cycles.
An {\bf ancestral graph} is said to be {\bf maximal} if every missing edge corresponds to a structural independence. %
If an ancestral graph $\mathcal{G}$ is not maximal, there exists a unique maximal ancestral graph $\Bar{\mathcal{G}}$ by adding bidirected edges to $\mathcal{G}$ \cite[]{richardson2002}.

\subsubsection{{\em ac}-connecting paths and {\em ac}-connected subsets}

Let us now define {\bf ancestor collider connecting paths} or {\bf {\em ac}-connecting paths}, which entail simpler path connecting criterion than the traditional {\bf m-connecting criterion}, discussed in the Appendix~A. Yet, {\bf {\em ac}-connecting paths}  and {\bf {\em ac}-connected subsets} will turn out to be directly relevant to characterize the likelihood decomposition and Markov equivalent classes of ancestral graphs.

\noindent
    {\bf Definition~2}.~[$ac$-connecting path] An $ac$-connecting path between $X$ and $Y$ given a subset of variables $\bm{C}$ (possibly including $X$ and $Y$) %
    is a collider path, 
$X\,*\!\!\rightarrow Z_1\longleftrightarrow \cdots \longleftrightarrow Z_K\leftarrow\!\!*\, Y$, with all  $Z_i\in \mathbf{An}_\mathcal{G}(\{X,Y\} \cup \bm{C})$, that is, with $Z_i$ in $\bm{C}$ or connected to $\{X,Y\} \cup \bm{C}$ by an ancestor path, {\it i.e.}~$Z_i\to\cdots\to T$ with $T\in \{X,Y\} \cup \bm{C}$.

\noindent
    {\bf Definition~3}.~[$ac$-connected subset] A subset $\bm{C}$ is said to be $ac$-connected  if $\forall X,Y\!\in\!\bm{C}$, $X$ and $Y$ are connected (through any type of edge) or there is an $ac$-connecting path between $X$ and $Y$ given~$\bm{C}$.

\subsection{Likelihood decomposition of ancestral graphs}

\vspace*{0.2cm}

\noindent
\mbox{{\bf Theorem~1.} [{\bf likelihood of ancestral graphs}]}
     {\it The cross-entropy $H(p,q)$ and likelihood $\mathcal{L}_{\mathcal{D\vert G}}$ of an ancestral graph $\mathcal{G}$ is decomposable in terms of multivariate cross-information, $I(\bm{C})$, summed over all $ac$-connected subsets of variables, $\bm{C}$ %
       (Definition~3),
\begin{eqnarray}
H(p,q)\!\!&=&\!\!-\!\!\!\!\sum_{\bm{C}\subseteq\bm{V}}^{\rm ac-connected}\!\!\!\!(-1)^{|\bm{C}|}I(\bm{C}) \nonumber\\
\mathcal{L}_{\mathcal{D}|\mathcal{G}}\!\!&=&\!\!{1\over Z_{\mathcal{D},\mathcal{G}}}\exp\Big(N\!\!\!\!\sum_{\bm{C}\subseteq\bm{V}}^{\rm ac-connected}\!\!\!\!(-1)^{|\bm{C}|}I(\bm{C})\Big)\hspace*{0.7cm}%
\label{LikelihoodDecomposition}
\end{eqnarray}
where $N$ is the number of iid samples in the dataset $\mathcal{D}$ and $Z_{\mathcal{D},\mathcal{G}}$ a data- and model-dependent normalization constant.
}

     The proof of Theorem~1 is left to Appendix~B. It is based on a partition of the cross-entropy (Eq.~\ref{MIE}) into cross-information contributions from $ac$-connected and non-$ac$-connected subsets of variables,
{\color{black}which does not rely on  head-and-tail factorizations nor on imset formalism\footnote{The genesis of Theorem~1 and Proposition~3 is further discussed in \url{https://openreview.net/forum?id=Z2f4Laqi8U&noteId=8GLWeaAKc9}.}
Hu and Evans \cite[]{hu2020} proposed an equivalent result (Proposition~3.3 in \cite[]{hu2020}) with a proof using head-and-tail decomposition to define parametrizing sets, which happen to coincide with the $ac$-connected sets defined here (Definition~3)}.
     Theorem~1 characterizes in particular the Markov equivalence class of ancestral graphs
     \cite[]{richardson2002,spirtes1996,richardson2003,ali2002,ali2005,tian2005,ali2009} as,

\noindent
{\bf Corollary~2.} {\it Two ancestral graphs are Markov equivalent if and only if they have the same $ac$-connected subsets of vertices.}

{Note, in particular, that Eq.~\ref{LikelihoodDecomposition} holds for {\em maximal ancestral graphs} (MAG), for which %
  all pairs of $ac$-connected variables are connected by an edge,
and their Markov equivalent representatives,  the {\em partial ancestral graphs} (PAG) \cite[]{richardson2002,richardson1999a,zhang2007,zhang2008}. 

\noindent
{\bf Proposition 3.} The likelihood decomposition of ancestral graphs (Eq.~\ref{LikelihoodDecomposition}, Theorem~1)
can be estimated by replacing the model distribution $q$ by the empirical distribution  $p$ in the retained multivariate cross-information terms $I(\bm{C})$
corresponding to all {\it ac}-connected subsets of variables, $\bm{C}$.

Hence, Proposition~3 amounts to estimating all relevant cross-information terms in the likelihood function with the corresponding multivariate information terms computed from the available data, while assuming %
by construction that the model distribution obeys all local and global conditional independences entailed by the ancestral graph.
The corresponding
factorization of the model distribution can be expressed in terms of empirical distribution, assuming positive distributions, see Appendix~C.

\begin{figure*}[h!]
\vspace*{-0.5cm}
\begin{center}
\includegraphics[width=1.03\linewidth]{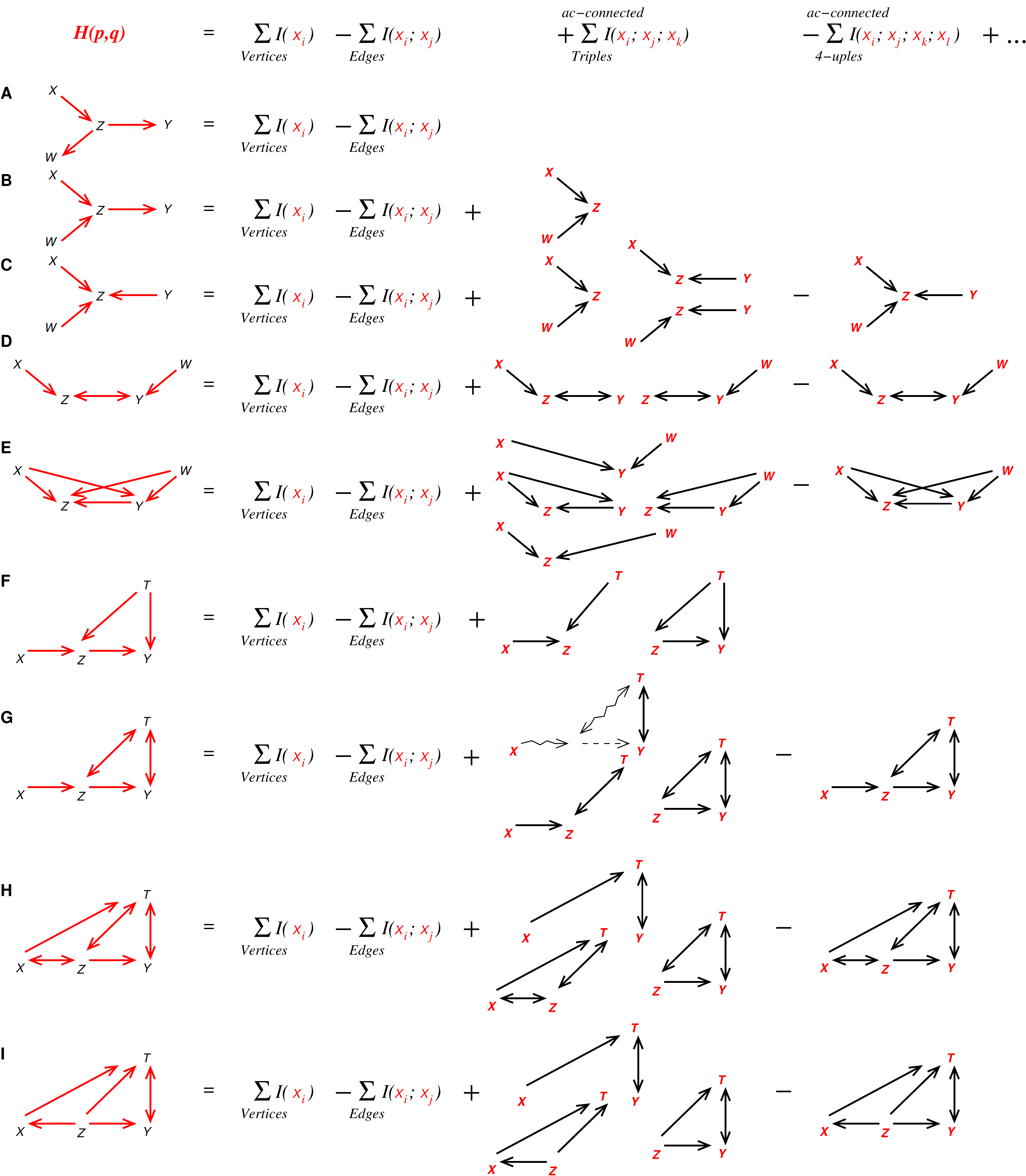}
\end{center}
\caption{\footnotesize{\bfseries Cross-entropy decomposition of ancestral graphs.}
 Examples of cross-entropy decomposition of ancestral graphs (red edges, lhs) in terms of relevant multivariate cross-information contributions 
 {\color{black}{$I(\bm{C})$ with $\bm{C}\subseteq\bm{V}$ (red nodes, rhs).
     Simple graphs: ({\sf\bfseries A}) without unshielded colliders, ({\sf\bfseries B}) with a single or non-overlapping unshielded colliders, ({\sf\bfseries C}) with overlapping unshielded colliders through three or more (conditionally) independent parents or  ({\sf\bfseries D}) through a two-(or more)-collider path. ({\sf\bfseries E}) Bayesian graph corresponding to the head-and-tail factorization of the two-collider path in ({\sf D}) estimated using the empirical distribution $p(.)$, see Appendix~C. ({\sf\bfseries F}) Simple Bayesian graph not Markov equivalent to an ancestral graph ({\sf\bfseries G})  sharing the same edges and unshielded collider \cite[]{ali2009}.  Solid black edges correspond to direct connections or collider paths
confined to the corresponding $ac$-connected subset  $\bm{C}$, while wiggly edges indicate collider paths extending beyond $\bm{C}$ yet indirectly connected to $\bm{C}$ by an ancestor path, marked with dashed edges, see Definition~2.
By contrast, graphs {\sf\bfseries H} and {\sf\bfseries I} illustrate the fact that collider paths may not be unique nor conserved between two Markov equivalent graphs ({\slshape i.e.}~sharing the same cross-information terms) \cite[]{ali2009}.}}}
\label{fig:CrossEntropyDecomposition_all}
\vspace*{-0.3cm}

\end{figure*}

Fig.~\ref{fig:CrossEntropyDecomposition_all} illustrates the
cross-entropy decomposition for a few graphical models 
in terms of cross-information contributions from their $ac$-connected subsets of vertices.
In particular, an unshielded non-collider ({\it e.g.}~$X\rightarrow Z\rightarrow W$, Fig.~\ref{fig:CrossEntropyDecomposition_all}A), is less likely ({\em i.e.}~higher cross-entropy) than %
an unshielded collider or `v-structure' ({\it e.g.}~$X\rightarrow Z\leftarrow W$, Fig.~\ref{fig:CrossEntropyDecomposition_all}B),
if the corresponding three-point information term is negative, $I(X;Z;W)<0$, in agreement with earlier results \cite[]{affeldt2015,verny2017}.
However, this early approach, exploiting the sign and magnitude of three-point information to orient v-structures, does not include higher order terms involving multiple v-structures, which can lead to orientation conflicts between unshielded triples, in practice.
Resolving such orientation conflicts requires to include information contributions from higher-order
$ac$-connected subgraphs, such as star-like $ac$-connected subsets including three or more parents, Fig.~\ref{fig:CrossEntropyDecomposition_all}C. Similarly, the cross-entropies of collider paths involving several colliders  also include higher-order terms, as with the simple example of a two-collider path, Fig.~\ref{fig:CrossEntropyDecomposition_all}D. 
By contrast, the cross-entropy based on the head-and-tail factorization of the same two-collider path, {\it i.e.}~$q(x,z,y,w)=q(z,y|x,w)q(x)q(w)$ \cite[]{richardson2009}, is found to be equivalent to the cross-entropy of a Bayesian graph without bidirected edge, Fig.~\ref{fig:CrossEntropyDecomposition_all}E, when estimated with the empirical distribution $p(.)$, see Appendix~C. This observation illustrates the difficulty to estimate the likelihood functions of ancestral graphs using head-and-tail factorization.

Further examples of graphical models, Figs.~\ref{fig:CrossEntropyDecomposition_all}F-I, 
show the relative simplicity of the decomposition with only few (non-trivial) {\color{black}{$ac$-connected}} contributing subsets $\bm{C}$ with \mbox{$|\bm{C}|\geqslant 3$}, as compared to the much larger number of  {\color{black}{non-$ac$-connected}} non-contributing subsets, that cancel each other by construction due to conditional independence constraints of the underlying model. 
Note, in particular, that most contributing multivariate information $I(\bm{C})$ only concern direct connections or collider paths within a single component subgraph induced by $\bm{C}$ (solid line edges {\color{black}{in Fig.~\ref{fig:CrossEntropyDecomposition_all}}}). However, occasionally, collider paths extending beyond $\bm{C}$ into $\mathbf{An}_{\mathcal{G}}(\bm{C})\setminus \bm{C}$ (marked with wiggly edges) with corresponding ancestor path(s) (marked with dashed edges) 
do occur, as shown in Fig.~\ref{fig:CrossEntropyDecomposition_all}G.

In addition, the present information-theoretic decomposition of the likelihood of ancestral graphs can readily distinguish their Markov equivalence classes according to Corollary~2.
For instance, the %
ancestral graphs of Fig.~\ref{fig:CrossEntropyDecomposition_all}F and Fig.~\ref{fig:CrossEntropyDecomposition_all}G, despite sharing %
the same edges and the same unshielded collider ($X \rightarrow Z \leftarrow T$), turn out not to be Markov equivalent, as discussed in \cite[]{ali2009}.
Indeed, %
their cross-entropy decompositions differ by two 
$ac$-connected contributing terms: a three-point cross information $I(X;Y;T)$ with a collider path  not confined in $\bm{C}$ 
({\em i.e.}~$X\rightsquigarrow Z \leftrightsquigarrow T \longleftrightarrow Y$ and corresponding ancestor path $Z \dashrightarrow Y$) and
 a four-point information term $I(X;Y;Z;T)$ due to the two-collider path ($X\rightarrow Z \longleftrightarrow T \longleftrightarrow Y$). 
More quantitatively, it shows that the graph of Fig.~\ref{fig:CrossEntropyDecomposition_all}G  with a two-collider path is more likely than the graph of Fig.~\ref{fig:CrossEntropyDecomposition_all}F whenever $I(X;Y;T)-I(X;Y;Z;T)=I(X;Y;T|Z)=I(X;Y|Z)-I(X;Y|Z,T)\!<\!0$. 
Finally, the Markov equivalent graphs of Fig.~\ref{fig:CrossEntropyDecomposition_all}H and Fig.~\ref{fig:CrossEntropyDecomposition_all}I, also due to \cite[]{ali2009}, illustrate the fact that the actual ancestor collider path between unconnected pairs does not need to be unique nor conserved between Markov equivalent graphs (as long as their cross-entropies share the same multivariate cross-information decomposition).

\section{Efficient search-and-score causal discovery using local information scores}

The likelihood estimation of ancestral graphs (Theorem 1 and Proposition 3) enables the implementation of a search-and-score algorithm for this broad class of graphs, which has attracted a number of contributions recently \cite[]{triantafillou2016,rantanen2021,claassen2022,andrews2022,hu2024a,hu2024b}. Our specific objective is not to develop an exact method limited to simple graphical models with a few nodes and small datasets but to implement an efficient and reliable heuristic method applicable to more challenging graphical models and large datasets.

Indeed,
search-and-score structure learning methods need to rely on heuristic rather than exhaustive search, in general,
given that the number of ancestral graphs grows super-exponentially as the number of vertices increases.
This can be implemented for instance with a Monte Carlo algorithmic scheme with random restarts, which efficiently probes relevant graphical models. Here, we opt, instead,  to use
the prediction of an efficient hybrid
causal discovery method, MIIC \cite[]{verny2017,cabeli2021,ribeiro-dantas2024}, as starting point for a subsequent %
search-and-score approach based on the proposed likelihood estimation of ancestral graphs (Eq.~\ref{LikelihoodDecomposition} and Proposition~3).

Moreover, while the likelihood decomposition of ancestral graphs may involve extended $ac$-connected subsets of variables, as illustrated in Fig.~\ref{fig:CrossEntropyDecomposition_all}, we aim to implement a computationally efficient search-and-score causal discovery method based on approximate local scores limited
to the close surrounding vertices of each node and edge.
Yet, while MIIC only relies on unshielded triple scores,
the novel search-and-score extension, MIIC\_search\&score,
uses also higher-order local information scores to compare alternative subgraphs, as detailed below.

The proposed method is shown to outperform %
MIIC and other state-of-the-art causal discovery methods on challenging datasets including latent variables.

\subsection{MIIC, an hybrid causal discovery method based on unshielded triple scores}

{\color{black}
  MIIC is an hybrid causal discovery method combining constraint-based and information-theoretic frameworks %
  \cite{verny2017,cabeli2020}.
  Unlike traditional constraint-based methods \cite[]{pearl_book2009,spirtes_book2000},
  MIIC does not directly attempt to uncover conditional independences but, instead, iteratively substracts the most significant three-point (conditional) information contributions of successive contributors, $A_1$, $A_2$, ..., $A_n$,
from the mutual information between each pair of variables, $I(X;Y)$, as,
\begin{equation}
I(X;Y) - I(X;Y;A_1) - I(X;Y;A_2|A_1) - \cdots - I(X;Y;A_n|\{A_i\}_{n-1})=I(X;Y|\{A_i\}_n)\;\;
\label{eq:3off2} 
\end{equation}
where $I(X;Y;A_k|\{A_i\}_{k-1}) > 0$ is the {\em positive} information contribution from $A_k$ to $I(X;Y)$ \cite{affeldt2015,affeldt2016}.
Conditional independence is eventually established when the residual conditional mutual information {\color{black}on the right hand side of Eq.~\ref{eq:3off2}}, $I(X;Y|\{A_i\}_n)$, becomes smaller than a complexity term, {\em i.e.}~$k_{X;Y|\{A_i\}}(N) \geqslant I(X;Y|\{A_i\}_n)\geqslant 0$,
which dependents on the considered variables and sample size $N$.

This leads to an undirected skeleton, which MIIC then (partially) orients based on the sign and amplitude of the regularized conditional 3-point information terms \cite{affeldt2015,verny2017}.
In particular, negative conditional 3-point information terms, %
$I(X;Y;Z|\{\!A_i\!\})\!<\!0$,
correspond to the signature of causality in observational data \cite{affeldt2015} and lead to the prediction of a v-structure, $X\rightarrow Z \leftarrow Y$, if $X$ and $Y$ are not connected in the skeleton. %
By contrast, a positive conditional 3-point information term, $I(X;Y;Z|\{\!A_i\!\})\!>\!0$, implies the absence of a v-structure and suggests to propagate the orientation of a previously directed edge $X\rightarrow Z \adjacent Y$ as $X\rightarrow Z \rightarrow Y$.

In practice, MIIC's strategy to circumvent spurious conditional independences significantly improves recall, that is, the fraction of correctly recovered edges, compared to traditional constraint-based methods \cite[]{affeldt2015,verny2017}.
Yet, MIIC only relies on unshielded triple scores to reliably uncover significant contributors and orient v-structures, as outlined above.
MIIC has been recently improved to ensure the consistency of the separating set in terms of indirect paths in the final skeleton or (partially) oriented graphs \cite[]{li2019,ribeiro-dantas2024} and to improve the reliably of predicted orientations \cite[]{cabeli2021,ribeiro-dantas2024}.

The predictions of this recent version of MIIC, which include three type of edges (directed, bidirected and undirected),  have been used as starting point for the subsequent local search-and-score method implemented in the present paper.

\subsection{New search-and-score method based on higher-order local information scores}

Starting from the structure predicted by MIIC, as detailed above,
MIIC\_search\&score  %
proceeds in {\color{black}two steps, first to remove likely %
  false positive edges (Step~1) and then to (re)orient the remaining edges based on their estimated contributions to the global likelihood decomposition, Eq.~\ref{LikelihoodDecomposition}  (Step~2).}

\subsubsection{Step 1: Node scores for edge orientation priming and edge removal}

The first step consists in minimizing a node score corresponding to the local normalized log likelihood of each node  w.r.t.~its possible parents or spouses amongst the connected nodes predicted by MIIC.
To this end, the node score assesses the conditional entropy of each node w.r.t.~a selection of %
{\color{black}parents, spouses or neighbors}, $\mathbf{Pa'}_{\!_{X_i}}\!\subseteq\mathbf{Pa}_{_{X_i}}\!\!\cup \mathbf{Sp}_{_{X_i}}\!\!\cup \mathbf{Ne}_{_{X_i}}\!$,
and a factorized Normalized Maximum Likelihood (fNML) regularization \cite[]{affeldt2015}, see Appendix~D for details,
\vspace*{-0.3cm}

\begin{equation}
{\rm Score_{n}}(X_i) = H(X_i|\mathbf{Pa'}_{\!_{X_i}}) + {1\over N}\sum_j^{q_{x_i}}\log\mathcal{C}^{r_{x_i}}_{n_j}\label{eq:nodescore}
\end{equation}
where $q_{x_i}$ corresponds to the combination of levels of $\mathbf{Pa'}_{\!_{X_i}}\!$, 
while  $r_{x_i}$ is the number of levels of $X_i$, and $n_j$ the number of samples corresponding to a particular combination of levels $j$ in each summand, with $\sum_j n_j = N$, the total number of samples. $\log\mathcal{C}^{r_{x_i}}_{n_j}$ is the fNML regulatization cost summed over all combinations of levels, %
$q_{x_i}$, \cite[]{kontkanen2007,roos2008}, see Appendix~D.

This first algorithm is looped over each node, priming the orientations of their surrounding edges  {(as directed, bidirected or undirected)}, until convergence. %
{Edges without orientation priming at either extremity are {\color{black}assumed to be false positive edges and} removed at the end of Step~1.}  

\subsubsection{Step 2: Edge orientation scores, as local contributions to the global likelihood score}

{\color{black}The second step consists in orienting the edges retained after Step~1, based on the optimization of
  their local contributions to the global likelihood score,
  Eq.~\ref{LikelihoodDecomposition}, %
  restricted to %
  $ac$-connected subsets containing up to two-collider paths.
  This amounts to minimizing each edge orientation score w.r.t.~its nodes' parents and spouses, corresponding to minus the conditional information plus a
fNML complexity cost, Table~1, 
given three sets of parents and spouses of $X$ and $Y$, {\slshape i.e.}~$\mathbf{Pa'}_{\!_{X\!\setminus\! Y}}\!=\mathbf{Pa}_{_{X}}\!\cup \mathbf{Sp}_{_{X}}\!\!\setminus\! Y$, ~$\mathbf{Pa'}_{\!_{Y\!\setminus\! X}}\!=\mathbf{Pa}_{_{Y}}\!\cup \mathbf{Sp}_{_{Y}}\!\!\setminus\! X$ and ~$\mathbf{Pa'}_{\!_{XY}}\!=\mathbf{Pa'}_{\!_{X\!\setminus\! Y}}\!\cup\mathbf{Pa'}_{\!_{Y\!\setminus\! X}}$ with their corresponding combinations of levels, $q_{_{y\!\setminus\! x}}$, $q_{_{x\!\setminus\! y}}$ and $q_{_{xy}}$.} These orientation scores, %
Table~1,
include symmetrized fNML complexity terms to enforce Markov equivalence, if  $X$ and $Y$ share the same parents or spouses (excluding $X$ and $Y$), see Appendix~D. Indeed, all three scores become equals if $\mathbf{Pa'}_{\!_{Y\!\setminus\! X}}=\mathbf{Pa'}_{\!_{X\!\setminus\! Y}}=\mathbf{Pa'}_{\!_{XY}}$ implying also the same combinations of parent and spouse levels,  %
$q_{_{y\!\setminus\! x}}=q_{_{x\!\setminus\! y}}=q_{_{xy}}$.

\vspace*{-0.3cm}

\begin{table}[h]
  \medskip
        \caption{{Local scores for the orientation of a single directed or bidirected edge, {\color{black}see Appendix~D}.}}  %
        \hrule
        \smallskip
     \begin{tabular}{lll}
        {\footnotesize Edge} & {\footnotesize Information} & {\footnotesize Symmetrized fNML complexity (Markov equivalent)}  \medskip\\
{\footnotesize $\!\!\!X\to Y$} & {\footnotesize $\!\!-I(X;Y|\mathbf{Pa'}_{\!_{Y\!\setminus\! X}})$} & {\footnotesize $\!\!\!{1\over 2N}\Big(\sum_j^{q_{_{x\!\setminus\! y}}\!r_y}\log \mathcal{C}_{n_j}^{r_x}-\sum_j^{q_{_{x\!\setminus\! y}}}\log \mathcal{C}_{n_j}^{r_x}+\sum_j^{q_{_{y\!\setminus\! x}}\!r_x}\log \mathcal{C}_{n_j}^{r_y}-\sum_j^{q_{_{y\!\setminus\! x}}}\log \mathcal{C}_{n_j}^{r_y}\Big)$}\medskip \\
{\footnotesize $\!\!\!X\leftarrow Y$} & {\footnotesize $\!\!-I(X;Y|\mathbf{Pa'}_{\!_{X\!\setminus\! Y}})$} & {\footnotesize $\!\!\!{1\over 2N}\Big(\sum_j^{q_{_{x\!\setminus\! y}}\!r_y}\log \mathcal{C}_{n_j}^{r_x}-\sum_j^{q_{_{x\!\setminus\! y}}}\log \mathcal{C}_{n_j}^{r_x}+\sum_j^{q_{_{y\!\setminus\! x}}\!r_x}\log \mathcal{C}_{n_j}^{r_y}-\sum_j^{q_{_{y\!\setminus\! x}}}\log \mathcal{C}_{n_j}^{r_y}\Big)$}  \medskip \\
{\footnotesize $\!\!\!X\leftrightarrow Y$} &  {\footnotesize $\!\!-I(X;Y|\mathbf{Pa'}_{\!_{XY}})$} & {\footnotesize $\!\!\!{1\over 2N}\Big(\sum_j^{q_{_{xy}}r_y}\log \mathcal{C}_{n_j}^{r_x}-\sum_j^{q_{_{xy}}}\log \mathcal{C}_{n_j}^{r_x}\,+\sum_j^{q_{_{yx}}r_x}\log \mathcal{C}_{n_j}^{r_y}-\sum_j^{q_{_{yx}}}\log \mathcal{C}_{n_j}^{r_y}\Big)$}\\
     \end{tabular}
     \smallskip
     \hrule
     \medskip
    \end{table}
\vspace*{-0.1cm}

{\color{black}While orientation scores cannot be summed over individual edges due to multiple countings of $ac$-connected contributions,
  score differences between alternative orientations provide an estimate of the global score change.
Hence, step~2}
algorithm is looped over each edge to compute an orientation score decrement, given {\color{black}its current orientation and} the orientations of surrounding edges. The orient\-ation change corresponding to the {\color{black}largest global
  score decrement, %
  without forming
  new directed or almost directed triangular cycles,} is then chosen at each iteration until convergence or until a limit cycle is reached.  {\color{black}Limit cycles may originate from the local two-collider approximation of the global score.} %

\section{Experimental results}
We first tested whether
MIIC\_search\&score orientation scores (Table~1) effectively predicts bidirected orientations on three simple ancestral models, Fig.~\ref{fig:toy} {\color{black}in Appendix~E}, when the end nodes do not share the same parents (Fig.~\ref{fig:toy}, Model~1), share some parents (Fig.~\ref{fig:toy}, Model~2) or when the bidirected edge is part of a longer than two-collider paths  (Fig.~\ref{fig:toy}, Model~3). The prediction of the edge orientation scores are summarized in Table~3, Appendix~E, and show good predictions for large enough datasets.

Beyond these simple examples, focussing on the discovery of bidirected edges in small toy models of ancestral graphs, we also analyzed more challenging benchmarks from the bnlearn repository \cite[]{scutari2010}, {\color{black}Figs.~\ref{fig:benchmarks1}-\ref{fig:benchmarks_bootstrap}}.
They concern ancestral graphs obtained by hiding up to 20\% of variables in {\color{black} Discrete} Bayesian Networks of increasing complexity (number of nodes and parameters), such as Alarm (37 nodes, 46 links, 509 parameters), Insurance (27 nodes, 52 links, 1,008 parameters), {\color{black}Barley (48 nodes, 84 links, 114,005 parameters), and Mildew (35 nodes, 46 links, 540,150 parameters)}. We then assessed causal discovery performance in terms of {\it Precision}, $TP/(TP+FP)$, and {\it Recall}, $TP/(TP+FN)$, relative to the theoretical PAGs, while counting as false positive ($FP$), all correctly predicted edges but %
with a different orientation as the directed or bidirected edges of the PAG.

 {\color{black}Figs.~\ref{fig:benchmarks1}-\ref{fig:benchmarks_bootstrap} compare} MIIC\_search\&score performance to MIIC results used as starting point for MIIC\_search\&score  and to FCI \cite[]{zheng2024}.
{\color{black}Fig.~\ref{fig:benchmarks1} results are obtained from independent datasets for each ancestral graph and sample size, while  Fig.~\ref{fig:benchmarks_bootstrap} results provide a bootstrap sensitivity analysis to sampling noise for each method based on independent resamplings with replacement of single datasets of increasing size.} %
MIIC and MIIC\_search\&score settings were set as described in section 3 above. The open-source MIIC R package (v1.5.2, GPL-3.0 license) was obtained at \url{https://github.com/miicTeam/miic_R_package}.
FCI from the python causal-learn package (v0.1.3.8, MIT license) \cite[]{zheng2024} was obtained at \url{https://github.com/py-why/causal-learn} and run with G$^2$-conditional independence test and default parameter $\alpha= 0.05$.

Overall, MIIC\_search\&score is found to outperform MIIC in terms of edge precision with little to no decrease in edge recall, {\color{black}Figs.~\ref{fig:benchmarks1}-\ref{fig:benchmarks_bootstrap}}, demonstrating the benefit of MIIC\_search\&score's rationale to improve MIIC predictions by extending MIIC information scores from unshielded triples to higher-order information contributions. These originate from $ac$-connected subsets including nodes with more than two parents or spouses, or $ac$-connected subsets including two-collider paths.

{\color{black}MIIC\_search\&score is also found to outperform FCI on both precision and recall on small datasets ({\it e.g.}~$N\leqslant 10,000$ samples) of complex graphical models ({\it i.e.}~Insurance, Barley and Mildew), while reaching similar performance at larger sample sizes or for simpler graphical models ({\it i.e.}~similar precision on Alarm model), as expected from the asymptotic consistency of FCI for very large datasets, Fig.~\ref{fig:benchmarks1}.
We also observed that FCI had a hard time to converge on bootstrapped datasets, explaining the lack of FCI comparison with MIIC and MIIC\_search\&score in Fig.~\ref{fig:benchmarks_bootstrap}, except for Alarm on all sample sizes tested ($N\leqslant 20,000$) and for Insurance at small sample sizes ($N\leqslant 1,000$).
}

\vspace*{0.3cm}

\begin{figure*}[h!]
\vspace*{-0.5cm}
\begin{center}
\includegraphics[width=\linewidth]{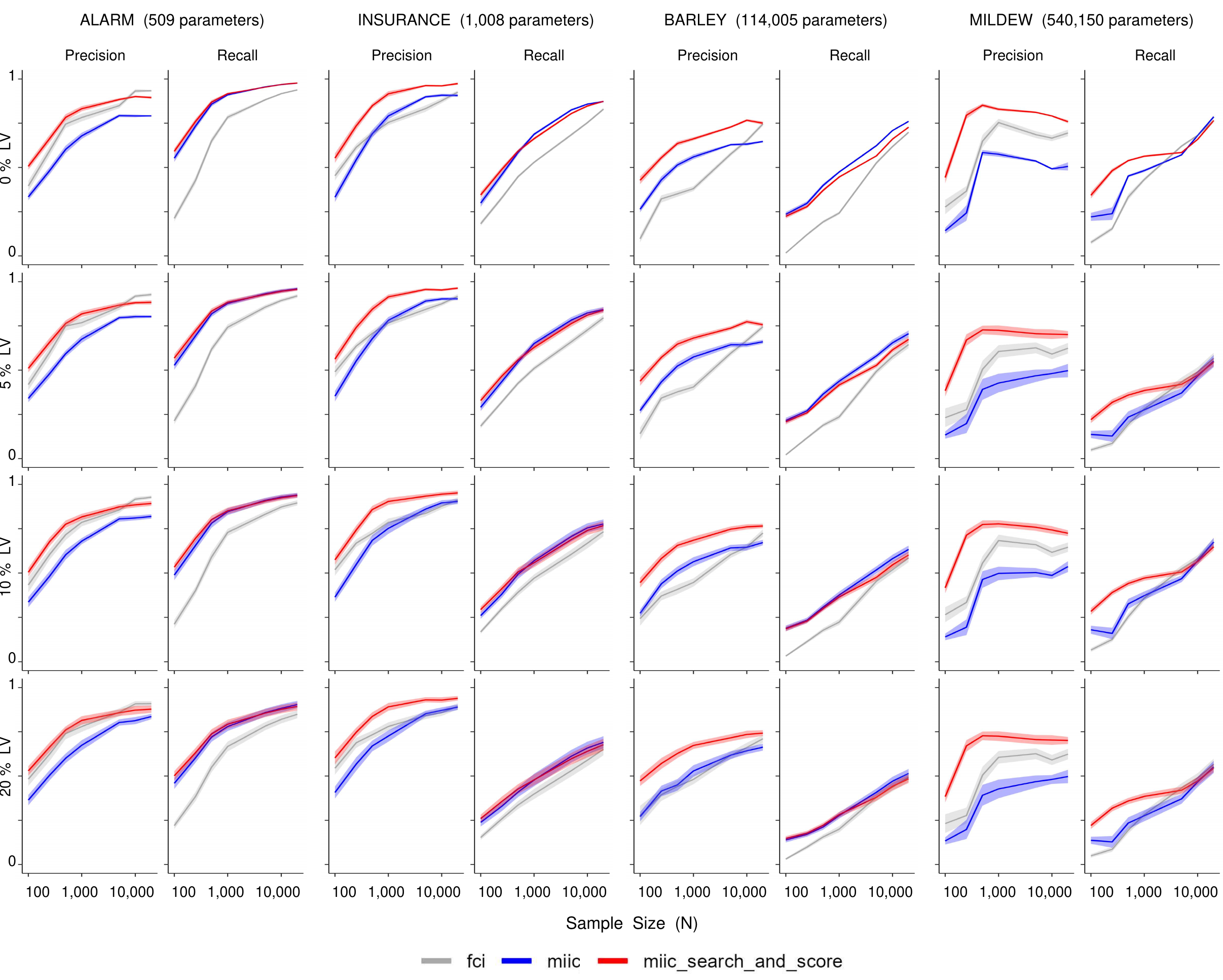}
\end{center}
\caption{\footnotesize{\bfseries Benchmark results on
    ancestral graphs of increasing complexity.}
  Benchmark results {\color{black}are averaged over 50 independent categorical datasets from}
  ancestral graphs obtained by hiding 0\%, 5\%, 10\% or 20\% of variables in {\color{black} Discrete} Bayesian Networks of increasing complexity (see main text): {\color{black}Alarm, Insurance, Barley and Mildew}. MIIC\_search\&score results are compared to MIIC results used as starting point for MIIC\_search\&score and FCI \cite[]{zheng2024}. Causal discovery performance is assessed in terms of {\it Precision} and {\it Recall} relative to the theoretical PAGs, while counting as false positive all correctly predicted edges but %
  with a different orientation as the directed or bidirected edges of the PAG. {\color{black}Error bars: 95\% confidence interval.}%
}
\label{fig:benchmarks1}
\end{figure*}

Importantly, the benchmark PAGs used to score the causal discovery results with increasing proportions of latent variables, {\color{black}Figs.~\ref{fig:benchmarks1}-\ref{fig:benchmarks_bootstrap}},
include not only bidirected edges originating from hidden common causes %
but also additional directed or undirected edges arising, in particular, from indirect effects of hidden variables with observed parents.
Irrespective of their orientations, all these additional edges originating from indirect effects of hidden variables generally correspond to weaker effects ({\it i.e.}~lower mutual information of indirect effects due to the Data Processing Inequality) and are more difficult to uncover than the edges of the original graphical model without hidden variables. {\color{black}This explains the steady decrease in recall for complex ancestral models ({\it i.e.}~Insurance, Barley, Mildew) with higher proportions of hidden variables, while precision remains essentially unaffected,  Figs.~\ref{fig:benchmarks1}-\ref{fig:benchmarks_bootstrap}.}

{\color{black}All in all, these results highlight MIIC\_search\&score capacity to efficiently and robustly learn complex graphical models from limited available data, which is a frequent setting for many real-world applications, in practice.
In addition, MIIC\_search\&score, which has been implemented to analyze challenging categorical datasets, is quite unique in this regard, as all other search-and-score methods for ancestral graphs  \cite[]{triantafillou2016,rantanen2021,claassen2022,andrews2022,hu2024b} have only been
  demonstrated
  on continuous datasets from linear Gaussian models and could not be included in the present benchmarks, Figs.~\ref{fig:benchmarks1}-\ref{fig:benchmarks_bootstrap}.
}

\vspace*{0.7cm}

\begin{figure*}[h!]
\vspace*{-0.5cm}
\begin{center}
  \includegraphics[width=\linewidth]{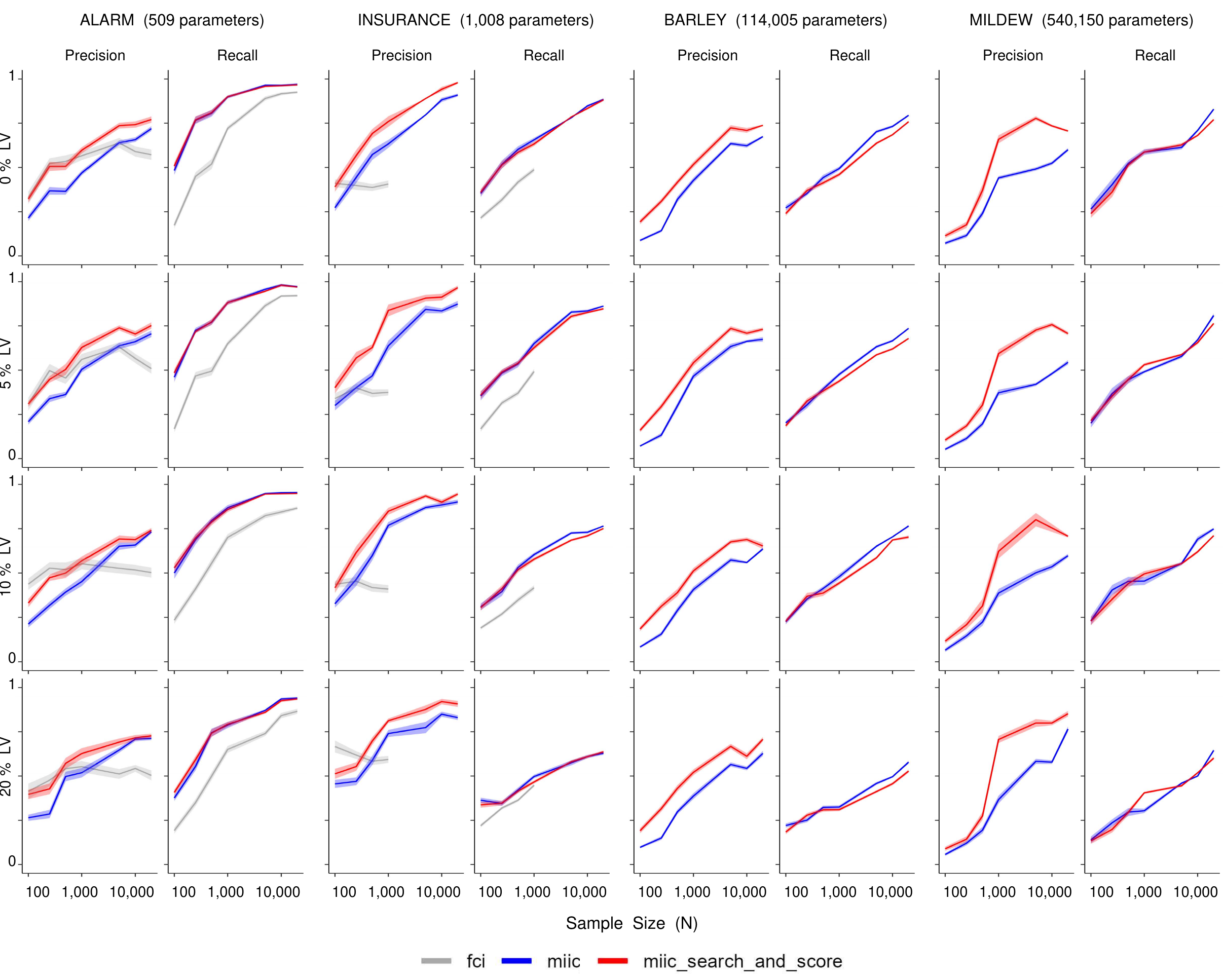}
\end{center}
\vspace*{-0.3cm}
\caption{\footnotesize{\color{black}\bfseries Benchmark results on bootstrap datasets from ancestral graphs of increasing complexity.}
  {\color{black}Benchmark results on bootstrap sensitivity analysis
to sampling noise based on 30 independent resamplings with replacement of single
datasets of increasing sizes. Ancestral graphs are obtained by hiding 0\%, 5\%, 10\% or 20\% of variables in {\color{black} Discrete} Bayesian Networks of increasing complexity (see main text): {\color{black}Alarm, Insurance, Barley and Mildew}. MIIC\_search\&score results are compared to MIIC results used as starting point for MIIC\_search\&score and FCI \cite[]{zheng2024}.
The lack of FCI results, except for Alarm on all sample sizes tested ($N\leqslant 20,000$) and for Insurance at small sample sizes ($N\leqslant 1,000$), stems from FCI difficulty to converge on bootstrapped datasets.
Causal discovery performance is assessed in terms of {\it Precision} and {\it Recall} relative to the theoretical PAGs, while counting as false positive all correctly predicted edges but %
with a different orientation as the directed or bidirected edges of the PAG. {\color{black}Error bars: 95\% confidence interval.}%
}}
\label{fig:benchmarks_bootstrap}
\vspace*{-0.3cm}

\end{figure*}

\section{Limitations}
The main limitation of the paper concerns the local scores used in the search-and-score algorithm, which are limited to $ac$-connected subsets of vertices with a maximum of two-collider paths.

While this approach could be extended to higher-order  information contributions including {\color{black}three-or-more-collider paths}, it allows for a simple two-step search-and-score scheme at the level of individual nodes (step~1) and edges (step~2), as detailed in section 3. This already shows a significant improvement in causal discovery performance ({\it i.e.}~combining good precision and good recall on challenging benchmarks) as compared to existing state-of-the-art methods. %

\begin{ack}
{\color{black}
HI  acknowledges support from CNRS, Institut Curie and Prairie Institute, as well as funding from ANR-22-PESN-0002 ``AI4scMed''  and ANR 23CE13001802 ``Patterning'' grants.
NL acknowledges a PhD fellowship from CNRS-Imperial College joint PhD programme.}
\end{ack}

%

%

%

%

%

%

%
%

%
%
%

%
%
%
%
%
%
%

%
%

%
%
%
%
%
%
%
%
%

%
%

%

%
%
%
%
%
%
%
%

%

%
%

%
%%%% Bibliography %%%%%%%%%%%%%%%%%%%%%%%%%%%%%%%%%%%%%%%%
%{\small
%\bibliographystyle{ScienceAdvances_wt-title}
%\bibliography{scibib}
%}

%\vspace*{0.8cm}

%
%
%
%
%

\newpage

\appendix

\section*{Appendix / supplemental material}

\section{Preliminaries: connection and separation criteria}

\subsection{{\em m}-connection {\em ~vs~} {\em m'}-connection criteria}

An ancestral graph can be interpreted as encoding a set of conditional indepencence relations by a graphical criterion, called $m$-separation, based on the concept of $m$-connecting paths, which generalizes the separation criteria of Markov and Bayesian networks to ancestral graphs.

\noindent
{\bf Definition~4.} [$m$-connecting path] A path $\bm{\pi}$ between $X$ and $Y$ is $m$-connecting given a (possibly empty) subset $\bm{C}\subseteq \bm{V}$ (with $X,Y\notin \bm{C}$) if:\\
\hspace*{1cm}{\em i)} ~its non-collider(s) are not in $\bm{C}$, and\\
\hspace*{1cm}{\em ii)} its collider(s) are in $\mathbf{An}_\mathcal{G}(\bm{C})$.

\noindent
{\bf Definition 5.} [$m$-separation criterion] The subsets $\bm{A}$ and $\bm{B}$ are said to be $m$-separated by $\bm{C}$, noted $\bm{A}\!\perp_{m}\!\bm{B|\bm{C}}$, if there is no $m$-connecting path between any vertex in  $\bm{A}$ and any vertex in $\bm{B}$ given $\bm{C}$.

The probabilistic interpretation of ancestral graph is given by its global and pairwise Markov properties (which are equivalent \cite[]{richardson2002}): if  $\bm{A}$ and $\bm{B}$ are $m$-separated by $\bm{C}$, then $\bm{A}$ and $\bm{B}$ are conditionally independent given $\bm{C}$ and $\forall X\in \bm{A}$ and  $\forall Y\in \bm{B}$, there is a probability distribution $P$ faithful to $\mathcal{G}$ such that their conditional mutual information vanishes, {\em i.e.}~$I_P(X;Y|\bm{C})=0$, also noted $X\perp\!\!\!\perp_P Y|\bm{C}$.

\noindent
However, as discussed above, the proof of Theorem~1 will require to introduce a weaker $m^\prime$-connection criterion defined below.

\noindent
{\bf Definition~6.} [$m^\prime$-connecting path] A path $\bm{\pi}$ between $X$ and $Y$ is $m^\prime$-connecting given a subset $\bm{C}\subseteq \bm{V}$ (with $X,Y$ possibly in $\bm{C}$) if:\\
\hspace*{1cm}{\em i)} ~its non-collider(s) are not in $\bm{C}$, and\\
\hspace*{1cm}{\em ii)} its collider(s) are in $\mathbf{An}_\mathcal{G}(\{X,Y\} \cup \bm{C})$.

Note, in particular, that an $m$-connecting path is necessary an $m^\prime$-connecting path but that the converse is not always true. For example, the path \mbox{$X\!\rightarrow\!Z\!\longleftrightarrow\!T\!\longleftrightarrow\!Y$} in Fig.~\ref{fig:CrossEntropyDecomposition_all}G (with $Z\rightarrow Y$) is an $m^\prime$-connecting path given $T$ (as $Z\in \mathbf{An}_\mathcal{G}(\{X,Y\} \cup T)$) but not an $m$-connecting path given $T$ (as $Z\notin \mathbf{An}_\mathcal{G}(T)$).

\noindent
However, Richardson and Spirtes 2002 \cite[]{richardson2002} have shown the following lemma,  

\noindent
{\bf Lemma~4.} [Corollary 3.15 in \cite[]{richardson2002}] {\it In an ancestral graph $\mathcal{G}$, there is a $m^\prime$-connecting  path $\bm{\mu}$ between $X$ and $Y$  given $\bm{C}$  if and only if there is a (possibly different) $m$-connecting path $\bm{\pi}$ between $X$ and $Y$ given $\bm{C}$.}

Hence, Lemma~4 implies that $m^\prime$-separation and $m$-separation criteria are in fact equivalent, as an absence of $m^\prime$-connecting paths implies an absence of $m$-connecting paths and vice versa. This enables to reformulate the $m$-separation criterion above as,

\noindent
{\bf Definition~7.} [$m^\prime$-separation (and $m$-separation) criteria] 
The subsets $\bm{A}$ and $\bm{B}$ are said to be $m^\prime$-separated (or $m$-separated) by $\bm{C}$, if all paths from any $X\in\bm{A}$ to any  $Y\in\bm{B}$ have either\\
\hspace*{1cm}{\em i)} ~a non-collider in $\bm{C}$, or\\
\hspace*{1cm}{\em ii)} a collider {\em not} in $\mathbf{An}_\mathcal{G}(\{X,Y\} \cup \bm{C})$.

The probabilistic interpretation of an ancestral graph is given by its (global) Markov property: if  $\bm{A}$ and $\bm{B}$ are $m$-separated (or $m^\prime$-separated) by $\bm{C}$, then $\bm{A}$ and $\bm{B}$ are conditionally independent given $\bm{C}$, noted as, $\bm{A}\perp_{m}\bm{B|\bm{C}}$.

\subsection{{\em ac}-connecting paths and {\em ac}-connected subsets}

Let us now recall the definition of {\bf ancestor collider connecting paths} or {\bf {\em ac}-connecting paths},
which is directly 
relevant to characterize the  likelihood decomposition and Markov equivalent classes of ancestral graphs (Theorem~1). We give here a different yet equivalent definition of {\em ac}-connecting paths as defined in the main text (Definition~2) in order to underline the similarities and differencies with the notion of $m^\prime$-connecting path (Definition~6).

\noindent
{\bf Definition~8.} [$ac$-connecting path] A path $\bm{\pi}$ between $X$ and $Y$ is {\color{black}{an $ac$-connecting path}} given a subset $\bm{C}\subseteq \bm{V}$ (with $X$ and $Y$ possibly in $\bm{C}$) if:\\
\hspace*{1cm}{\em i)} ~$\bm{\pi}$ does not have any noncollider, and\\
\hspace*{1cm}{\em ii)} its collider(s) are in $\mathbf{An}_\mathcal{G}(\{X,Y\} \cup \bm{C})$.

\noindent
Hence, more simply (following Definition~2 in the main text), 
an $ac$-connecting path given $\bm{C}$
is a collider path, 
$X\,*\!\!\rightarrow Z_1\leftrightarrow \cdots \leftrightarrow Z_K\leftarrow\!\!*\, Y$, with all  $Z_i\in \mathbf{An}_\mathcal{G}(\{X,Y\} \cup \bm{C})$, {\em i.e.}~with $Z_i$ in $\bm{C}$ or connected to $\{X,Y\} \cup \bm{C}$ by an ancestor path, $Z_i\to\cdots\to T$ with $T\in \{X,Y\} \cup \bm{C}$.

\noindent
{\bf Definition~9.} [$ac$-separation criterion] The subsets $\bm{A}$ and $\bm{B}$ are said to be $ac$-separated by $\bm{C}$ if there is no $ac$-connecting path between any vertex in  $\bm{A}$ and any vertex in $\bm{B}$ given $\bm{C}$.

\noindent
Previous definitions and Lemma~4 readily lead to the following corollary between the different connection and separation criteria:

\noindent
{\bf Corollary~5.} \\
{\it
\hspace*{0.41cm}{i)} \hspace*{0.2cm}$m$-connecting path $\pi$ ~$\Longrightarrow$~ $m^\prime$-connecting path $\pi$\\
\hspace*{0.41cm}{ii)} ~$ac$-connecting path $\pi$ ~$\Longrightarrow$~ $m^\prime$-connecting path $\pi$\\
\hspace*{0.41cm}{iii)} \hspace*{1cm}{$m$-separation ~$\Longleftrightarrow$~ $m^\prime$-separation}\\
\hspace*{0.41cm}{iv)} \hspace*{0.5cm}{$m/m^\prime$-separation  ~$\Longrightarrow$~ $ac$-separation}
}

\noindent
Finally, we recall the notion of {\bf $\bm{ac}$-connected subset} (Definition~3 in the main text), which is central for the decomposition of the likelihood of ancestral graphs (Theorem~1): A subset $\bm{C}$ is said to be $ac$-connected  if $\forall X,Y\in \bm{C}$, there is an $ac$-connecting path between $X$ and $Y$ w.r.t.~$\bm{C}$.

\section{Proof of Theorem~1.}

In order to prove that the likelihood function of an ancestral graph, Eq.~\ref{LikelihoodDecomposition}, contains all and only the $ac$-connected subsets of vertices in $\mathcal{G}$ (Definition~3), we will first show %
({\em\bfseries i}) that all non-$ac$-connected subsets $\bm{S^\prime}$ are included in a cancelling combination of
multivariate {\color{black}cross-information terms,  $I(X;Y|\bm{A})=0$, with $X,Y\in \bm{S^\prime}$ and $\bm{S^\prime}\subseteq \bm{S}=\{X,Y\}\cup\bm{A}$, Eq.~\ref{CMIE}}.
Conversely, we will then show ({\em\bfseries ii}) that cancelling combinations of multivariate {\color{black}cross-information} terms associated to pairwise conditional independence, \mbox{$I(X;Y|\bm{A})=\sum_{\bm{S^\prime}\subseteq \bm{S}}^{X,Y\in \bm{S^\prime}}(-1)^{|\bm{S^\prime}|}I(\bm{S^\prime})=0$}, do not contain any $ac$-connected subset $\bm{S^\prime}$.
Finally, we will prove ({\em\bfseries iii}) that the information terms which appear in multiple cancelling combinations from different pairwise independence constraints %
do not modify the multivariate information decomposition of the likelihood function of ancestral graphs, Eq.~\ref{LikelihoodDecomposition}, as these shared/overlapping terms in fact all cancel through more global Markov independence relationships involving higher order (three or more points) vanishing multivariate information terms, such as $I(X;Y;Z|\bm{A})=0$.

{\em\bfseries i)} Let's first prove that all non-$ac$-connected subsets $\bm{S^\prime}$ are included in at least one cancelling combination of multivariate {\color{black}cross-information},  $I(X;Y|\bm{A})=0$, with $X,Y\!\in\!\bm{S^\prime}$ and $\bm{S^\prime}\!\subseteq\!\{\!X,Y\!\}\cup\bm{A}$.

If  $\bm{S^\prime}$ is a non-$ac$-connected subset, there is at least one disconnected pair $X$ and $Y$ for which each path $\pi_j$ between  $X$ and $Y$
contains either some collider(s) not in $\mathbf{An}_\mathcal{G}(\bm{S^\prime})$ or, if all colliders along  $\pi_j$ are in $\mathbf{An}_\mathcal{G}(\bm{S^\prime})$, there must be some non-collider(s) at node(s) $\bm{Z}_j$
but not necessarily in $\bm{S^\prime}$. Let's define $\bm{S}=\bm{S^\prime} \cup_j \bm{Z}_j$.
$X$ and $Y$ can be shown to be $m$-separated given $\bm{S}\setminus \{X,Y\}$,
as for each path  $\pi_j$ between $X$ and $Y$, its non-collider(s) are in $\bm{S}$  at node(s) $\bm{Z}_j$ (when all collider(s) along  $\pi_j$ are in $\bm{S^\prime}$) or there is some collider(s) not in $\mathbf{An}_\mathcal{G}(\bm{S^\prime})$,
which are not in $\mathbf{An}_\mathcal{G}({\color{black}\bm{S}})$ either.
The latter statement is proven by contradiction assuming that there is a collider at  $Z\notin \mathbf{An}_\mathcal{G}(\bm{S^\prime})$ such that $Z\in  \mathbf{An}_\mathcal{G}(\bm{S})$.  There is therefore  a directed path $Z\to \cdots \to W$ with $W\in \bm{S}$. Hence,  $W\in \bm{S^\prime}$ or there is a noncollider at $W\in  \bm{Z}_j$ which is on a path $\pi_j$ between $X$ and $Y$  along which all colliders  are in $\mathbf{An}_\mathcal{G}(\bm{S^\prime})$ by construction of $\bm{S}$.
This leads by induction %
to $Z\to \cdots \to W \to \cdots \to T$ where $T\in \bm{S^\prime}$ and thus  $Z\in \mathbf{An}_\mathcal{G}(\bm{S^\prime})$, which is a contradiction. 
Hence,  all non-$ac$-connected subsets $\bm{S^\prime}$ are included in a cancelling combination of multivariate {\color{black}cross-}information terms,  $I(X;Y|\bm{A})=0$, with $X,Y\in \bm{S^\prime}$ and $\bm{S^\prime}\subseteq \bm{S}=\{X,Y\}\cup\bm{A}$. %

{\em\bfseries ii)} Conversely, we will now show that cancelling combinations of multivariate {\color{black}cross-}information terms associated to pairwise conditional independence, \mbox{$I(X;Y|\bm{A})=\sum_{\bm{S^\prime}\subseteq \bm{S}}^{X,Y\in \bm{S^\prime}}(-1)^{|\bm{S^\prime}|}I(\bm{S^\prime})=0$}, %
do not contain any $ac$-connected subset $\bm{S^\prime}${\color{black}, where $\bm{S}=\{X,Y\}\cup\bm{A}$}.

We will prove it by contradiction assuming that  there exists a subset $\bm{W} \subseteq \bm{A}$, such that $\bm{S^\prime}=\{X,Y\}\cup\bm{W}$ is $ac$-connected.
In particular, there should be an $ac$-connecting path between $X$ and $Y$ confined to $\mathbf{An}_\mathcal{G}(\bm{S^\prime})$ and thus to $\mathbf{An}_\mathcal{G}(\bm{S})\supseteq \mathbf{An}_\mathcal{G}(\bm{S^\prime})$, which is an $m^\prime$-connecting path between $X$ and $Y$ given $\bm{A}$, %
contradicting the {\color{black}{above hypothesis of $m^\prime$-separation given $\bm{A}$, {\em i.e.}~$I(X;Y|\bm{A})=0$}}. The use of $m^\prime$-separation, {\em i.e.}~the absence of $m^\prime$-connecting paths with colliders in $\mathbf{An}_\mathcal{G}(\bm{S})$ rather than $m$-connecting paths with colliders in $\mathbf{An}_\mathcal{G}(\bm{A})$, is necessary here, see Definitions 4 and 6. Hence, no $ac$-connected subset $\bm{S^\prime}$ is included in cancelling combinations of multivariate {\color{black}cross-}information terms associated to pairwise conditional independence, \mbox{$I(X;Y|\bm{A})=\sum_{\bm{S^\prime}\subseteq \bm{S}}^{X,Y\in \bm{S^\prime}}(-1)^{|\bm{S^\prime}|}I(\bm{S^\prime})=0$}. %

{\em\bfseries iii)} Finally, we will show that the information terms which appear in multiple cancelling combinations from different pairwise independence constraints do not modify the multivariate {\color{black}cross-}information decomposition of the likelihood function of ancestral graphs, Eq.~\ref{LikelihoodDecomposition}, as these shared/overlapping terms in fact all cancel through more global Markov independence relationships involving higher order (three or more points) vanishing multivariate {\color{black}cross-}information terms, such as $I(X;Y;Z|\bm{A})=0$.

This result requires to use an ordering of the nodes,  $X_k\succ X_j \succ X_i$, that is compatible with the directed edges of the %
ancestral graph
assumed to have no undirected edges, {\it i.e.}~$X_j \notin \mathbf{An}(X_i)$ if $X_j \succ X_i$. Under this ordering,  higher order nodes {\color{black}$X_k\succ X_j \succ X_i$} can be a priori excluded from all separating sets $\bm{A}_{ij}$ of pairs of lower order nodes, {\it i.e.}~if $I(X_i;X_j|\bm{A}_{ij})=0$ then $X_k\notin \bm{A}_{ij}$.

In particular, the two pairwise conditional independence relations $I(X_k;X_\ell|\bm{A}_{k\ell})=0$, with $X_\ell\succ X_k$, and  $I(X_i;X_j|\bm{A}_{ij})=0$, with $X_j \succ X_i$, do not share any multivariate {\color{black}cross-}information  terms, if $X_\ell \neq X_j$.
{\color{black}Indeed, as $I(X_i;X_j|\bm{A}_{ij})$  contains all {\color{black}cross-}information  terms including both $X_i$ and $X_j$ as well as every subset (possibly empty) of $\bm{A}_{ij}$, none of them includes $X_\ell$ if $X_\ell\succ X_j$. Therefore $I(X_i;X_j|\bm{A}_{ij})$ does not contain any {\color{black}cross-}information  term of $I(X_k;X_\ell|\bm{A}_{k\ell})$ which contains both $X_k$ and $X_\ell$ as well as every subset (possibly empty) of $\bm{A}_{k\ell}$.}
This property eliminates all multiple {\color{black}countings of multivariate {\color{black}cross-}information terms} %
if $X_\ell \neq X_j$. Note that this result does not hold in general for ancestral graphs including undirected edges.

Hence, the issue of redundant multivariate {\color{black}cross-}information  terms in the likelihood decomposition, Eq.~\ref{LikelihoodDecomposition}, is related to the conditional independences of two or more pairs, $\{X_i,X_r\}$, $\{X_j,X_r\}$, ..., $\{X_\ell,X_r\}$, sharing the same higher order node, {\color{black}$X_r$, {\it i.e.},~$I(X_k;X_r|\bm{A}_{kr})=0$ for $k=i,j,\cdots,\ell$.}
However, this situation also entails a more global Markov independence constraint between $X_r$ and $\{X_i,X_j,\cdots,X_\ell\}$, given a separating set $\bm{A}$, {\color{black} with $\bm{A}_{kr}\subseteq \bm{A}\cup\{X_i,\cdots,X_\ell\}$  for $k=i,j,\cdots,\ell$. Such a global Markov independence constraint} can be decomposed into more local independence constraints using the chain rule {\color{black}(in {\it any} order of the variables $X_i,X_j,\cdots,X_\ell$)} and the decomposition rules of multivariate {\color{black}(cross)} information (Eq.~\ref{DecompositionRule2}),
\begin{eqnarray}
\;0&\!=\!&I(\{X_i,X_j,\cdots,X_\ell\};X_r|\bm{A})\nonumber\\
&\!=\!&\big(I(X_i;X_r|\bm{A})+I(X_j;X_r|\bm{A},X_i)\big)+\big[I(X_k;X_r|\bm{A},X_i,X_j)\big]+\cdots+I(X_\ell;X_r|\bm{A},\cdots)\nonumber\\
&\!=\!&\big(I(X_i;X_r|\bm{A})+I(X_j;X_r|\bm{A})-I(X_i;X_j;X_r|\bm{A})\big)\nonumber\\
&&+\big[I(X_k;X_r|\bm{A},X_i)-I(X_j;X_k;X_r|\bm{A},X_i)\big]+\cdots+I(X_\ell;X_r|\bm{A},\cdots)\nonumber\\
&\!=\!&\big(I(X_i;X_r|\bm{A})+I(X_j;X_r|\bm{A})-I(X_i;X_j;X_r|\bm{A})\big)\nonumber\\
&&+\big[I(X_k;X_r|\bm{A})-I(X_j;X_k;X_r|\bm{A})-I(X_i;X_k;X_r|\bm{A})+I(X_i;X_j;X_k;X_r|\bm{A})\big]+\cdots\nonumber\hspace*{.5cm}%
\end{eqnarray}
where all the conditional multivariate {\color{black}cross-}information terms vanish by induction due to the non-negativity of (conditional) mutual {\color{black}(cross)} information. In particular, the conditional multivariate {\color{black}cross-}information terms in the last expression, {\it i.e.}~between $X_r$ and each subset of $\{X_i,X_j,\cdots,X_\ell\}$ given the separating set $\bm{A}$, all vanish.
This result can be readily extended to any subsets $\{X_r,X_s,\cdots,X_z\}$ (conditionally) independent of $\{X_i,X_j,\cdots,X_\ell\}$ given a separating set $\bm{A}$, {\em i.e.}~$I(\{X_i,X_j,\cdots,X_\ell\};\{X_r,X_s,\cdots,X_z\}|\bm{A})=0$.
Hence, as the final conditional multivariate cross-information terms of the decomposition all vanish while not sharing any subsets of variables, it proves the absence of redundancy and a global cancellation of non-$ac$-connected subsets (from pairwise and higher order conditional independence relations) in the likelihood function of ancestral graphs without undirected edges, Eq.~\ref{LikelihoodDecomposition}. %

Hence, only $ac$-connected subsets effectively contribute to the cross-entropy of an ancestral graph with only directed and bidirected edges, Eq.~\ref{LikelihoodDecomposition}.
\hfill$\square$

\newpage

\section{Factorization of the probability distribution of ancestral graphs}

\subsection{Factorization resulting from Theorem~1 and Proposition~3}

Before presenting the factorization of the model distribution of ancestral graphs resulting from Theorem~1 and Proposition~3, it is instructive to obtain an equivalent factorization for Bayesian graphs, assuming a positive empirical distributions, $p(x_1,\cdots,x_m)= \prod_{i=1}^{m} p(x_i|x_{i-1},\cdots,x_1)>0$,
\begin{eqnarray}
  q(x_1,\cdots,x_m)&=&\prod_{i=1}^{m} q(x_i|\mathbf{pa}_{x_i}) = \prod_{i=1}^{m} p(x_i|\mathbf{pa}_{x_i})\nonumber\\
  &=& p(x_1,\cdots,x_m) \prod_{i=1}^{m} {p(x_i|\mathbf{pa}_{x_i}) \over p(x_i|x_{i-1},\cdots,x_1)}\nonumber\\
  &=& p(x_1,\cdots,x_m) \prod_{i=1}^{m} {p(x_i|\mathbf{pa}_{x_i}) p(\bm{x}_{i-1}\!\!\setminus\!\mathbf{pa}_{x_i}|\mathbf{pa}_{x_i}) \over p(x_i,\bm{x}_{i-1}\!\!\setminus\!\mathbf{pa}_{x_i}|\mathbf{pa}_{x_i})} \label{eq:BNfactorization1}
\end{eqnarray}
This leads to the following alternative expressions for the cross-entropy $H(p,q)=-\sum_{\bm{x}}p(\bm{x})\log q(\bm{x})$ in terms of multivariate entropy and information, which only depend on the empirical joint distribution $p(\bm{x})$,
\begin{eqnarray}
  H(p,q)&=&\sum_{i=1}^{m} H(x_i|\mathbf{Pa}_{X_i})\nonumber\\
        &=&H(X_1,\cdots,X_m)+\sum_{i=1}^{m} I(X_i;\bm{X}_{i-1}\!\!\setminus\!\mathbf{Pa}_{X_i}|\mathbf{Pa}_{X_i})\label{BNfactorization2}
\end{eqnarray}
where $\sum_{i=1}^{m} I(X_i;\bm{X}_{i-1}\!\!\setminus\!\mathbf{Pa}_{X_i}|\mathbf{Pa}_{X_i})$ can be decomposed, using the chain rule and Eq.~\ref{CMIE}, into unconditional multivariate information terms,
which exactly cancel all the multivariate information of the non-$ac$-connected subsets of variables in the multivariate entropy decomposition, Eq.~\ref{MIE}.

Note, however, that this result obtained for Bayesian networks requires an explicit factorization of the global model distribution, $q(\bm{x})$, in terms of the empirical distribution, $p(\bm{x})$, which is not known and presumably does not exist, in general, for ancestral graphs.

Alternatively, assuming that the empirical and model distributions are positive ($\forall \bm{x}, p(\bm{x})>0$, $q(\bm{x})>0$), it is always possible to factorize them into factors associated to each (cross) information term in the (cross) entropy decomposition, %
Eq.~\ref{MIE}, as,
\begin{eqnarray}
  q(\bm{x})&=&\prod_{i=1}^{m} q(x_i) \times \prod_{i<j}^{m} {q(x_i,x_j)\over q(x_i)q(x_j)} \times \prod_{i<j<k}^{m} {q(x_i,x_j,x_k)q(x_i)q(x_j)q(x_k)\over q(x_i,x_j) q(x_i,x_k) q(x_j,x_k)}  \times\cdots~~~~~~~~\label{eq:qfactorization}
\end{eqnarray}
where all the marginal distributions over a subset of variables, {\it e.g.}~$q(x_i,x_j,x_k)=\sum_{\ell\neq i,j,k}q(\bm{x})$ or $p(x_i,x_j,x_k)=\sum_{\ell\neq i,j,k}p(\bm{x})$, cancel two-by-two by construction.

This can be illustrated on a simple example of a %
two-collider path
including one bidirected edge, $X\rightarrow Z \longleftrightarrow Y \leftarrow W$ (Fig.~\ref{fig:CrossEntropyDecomposition_all}D), valid for $q(.)$ and $p(.)$ alike,
\begin{eqnarray}
  q(x,z,y,w)\!\!&=&\!\!q(x)\;q(z)\;q(y)\;q(w)\nonumber\\
  \!\!&&\!\! \times\; {q(x,z)\over q(x)\,q(z)}\; {q(z,y)\over q(z)\,q(y)}\; {q(y,w)\over q(y)\,q(w)}\; {\color{lightgray}{q(x,y)\over q(x)\,q(y)}\; {q(x,w)\over q(x)\,q(w)}\; {q(z,w)\over q(z)\,q(w)}}\nonumber\\
  \!\!&&\!\!  \times\; {q(x)\,q(z)\,q(y)\,q(x,z,y)\over q(x,z)\,q(x,y)\,q(z,y)}\; {q(z)\,q(y)\,q(w)\,q(z,y,w)\over q(z,y)\,q(z,w)\,q(y,w)}\nonumber\\
  \!\!&&\!\!  \times\;  {\color{lightgray}{q(x)\,q(z)\,q(w)\,q(x,z,w)\over q(x,z)\,q(x,w)\,q(z,w)}\; {q(x)\,q(y)\,q(w)\,q(x,y,w)\over q(x,y)\,q(x,w)\,q(y,w)}}\nonumber\\
  \!\!&&\!\!  \times\; {q(x,z)\,q(z,y)\,q(y,w)\,q(x,y)\,q(x,w)\,q(z,w)\,q(x,z,y,w) \over q(x,z,y)\,q(x,z,w)\,q(x,y,w)\,q(z,y,w)\,q(x)\,q(y)\,q(z)\,q(w)}\label{fig:4nodefactorization}
\end{eqnarray}

\noindent
where all individual distribution marginals on subsets of variables, {\it e.g.}~$q(x)$, $q(x,z)$, $q(x,z,y)$ (or $p(x)$, $p(x,z)$, $p(x,z,y)$), cancel two-by-two by construction, except $q(x,z,y,w)$ (or $p(x,z,y,w)$).

In addition and {\it only for the model distribution} $q(.)$, all ratios in gray in Eq.~\ref{fig:4nodefactorization} also cancel due to Markov independence relations across non-$ac$-connected subsets (see proof of Theorem~1). This leaves a truncated factorization retaining all and only the $ac$-connected subsets of variables in the graph, which we propose to estimate on empirical data by substituting the remaining $q(.)$ terms by their empirical counterparts $p(.)$, see Proposition~3.

This leads to the following global factorization for $q(.)$ in terms of $p(.)$,
\begin{eqnarray}
  q(x,z,y,w)\!\!&\equiv&\!\!p(x)\;p(z)\;p(y)\;p(w) \; {p(x,z)\over p(x)\,p(z)}\; {p(z,y)\over p(z)\,p(y)}\; {p(y,w)\over p(y)\,p(w)} \nonumber\\
  \!\!&&\!\!  \times\; {p(x)\,p(z)\,p(y)\,p(x,z,y)\over p(x,z)\,p(x,y)\,p(z,y)}\; {p(z)\,p(y)\,p(w)\,p(z,y,w)\over p(z,y)\,p(z,w)\,p(y,w)} \nonumber\\
  \!\!&&\!\!   \times\; {p(x,z)\,p(z,y)\,p(y,w)\,p(x,y)\,p(x,w)\,p(z,w)\,p(x,z,y,w) \over p(x,z,y)\,p(x,z,w)\,p(x,y,w)\,p(z,y,w)\,p(x)\,p(y)\,p(z)\,p(w) }\nonumber\\
  \!\!&=&\!\! p(x,z,y,w)\;{\color{lightgray}{{p(x)\,p(y)\over  p(x,y)}\;
    {p(x)\,p(w)\over  p(x,w)}\;{p(z)\,p(w)\over  p(z,w)}}} \nonumber\\
  \!\!&&\!\!  {\color{lightgray}\times\;  {{p(x,z)\,p(x,w)\,p(z,w)\over p(x)\,p(z)\,p(w)\,p(x,z,w)}\; {p(x,y)\,p(x,w)\,p(y,w)\over p(x)\,p(y)\,p(w)\,p(x,y,w)}}} \label{fig:4nodefactorization2}
\end{eqnarray}

\noindent
where the terms in gray have been passed to the lhs of Eq.~\ref{fig:4nodefactorization}} applied to $p(.)$. This ultimately leads to the analog of the Bayesian Network factorization in Eq.~\ref{eq:BNfactorization1} but for the
two-collider path,
  $X\rightarrow Z \longleftrightarrow Y \leftarrow W$ (Fig.~\ref{fig:CrossEntropyDecomposition_all}D), 
\begin{eqnarray}
  q(x,z,y,w)\!\!&\equiv&\!\! p(x,z,y,w)\;{{p(x)\,p(w)\over  p(x,w)}} \; {{p(z|x)\,p(w|x)\over p(z,w|x)}}  \; {{p(x|w)\,p(y|w)\over p(x,y|w)}} \label{fig:4nodefactorization3}
\end{eqnarray}

\noindent
where the last three factors ``correct'' the expression of $p(x,z,y,w)$ for the three (conditional) independences entailed by the underlying graph, that is, $X\perp W$,  $Z\perp W|X$, and  $X\perp Y|W$.
\subsection{Relation to the head-and-tail factorizations}

The head-and-tail factorizations of the model distribution of an acyclic directed mixed graph, introduced by Richardson 2009 \cite[]{richardson2009}, {\color{black}do not correspond to a single factorized equation (as with Bayesian graphs, Eq.~\ref{eq:BNfactorization1}) but to multiple factorized equations, which} enable the parametrization of the joint probability distribution with independent parameters for ancestrally closed subsets of vertices.

For instance, the head-and-tail factorizations of the simple
two-collider path
including one bidirected edge, $X\rightarrow Z \longleftrightarrow Y \leftarrow W$, introduced above, Fig.~\ref{fig:CrossEntropyDecomposition_all}D, {\color{black}correspond to the following equations} \cite[]{richardson2009},
\begin{eqnarray}
  q(x,w)\!\!&=&\!\! {{q(x)\,q(w)}}\nonumber\\
  q(x,z)\!\!&=&\!\! q(z|x)\;{{q(x)}}\nonumber\\
  q(y,w)\!\!&=&\!\! q(y|w)\;{{q(w)}}\nonumber\\
  q(x,z,w)\!\!&=&\!\! q(z|x)\;{{q(x)\,q(w)}}\nonumber\\
  q(x,y,w)\!\!&=&\!\! q(y|w)\;{{q(w)\,q(x)}}\nonumber\\
  q(x,z,y,w)\!\!&=&\!\! q(z,y|x,w)\;{{q(x)\,q(w)}}
 \label{eq:4nodeheadandtailfactorization}
\end{eqnarray}
Importantly, these head-and-tail factorizations imply additional relations such as $q(y|w)=q(y|x,w)$ ({\it i.e.}~$X\perp Y|W$) obtained by comparing the last two relations in Eqs.~\ref{eq:4nodeheadandtailfactorization} after marginalizing $q(x,z,y,w)$ over $z$. However, such implicit conditional independence relations are {\it not verified by the empirical distribution $p(.)$ in general} and prevent the estimation of the head-and-tail factorizations by substituting the rhs $q(.)$ terms in Eqs.~\ref{eq:4nodeheadandtailfactorization} with their empirical counterparts $p(.)$, as in the case of Bayesian networks, Eq.~\ref{eq:BNfactorization1}.

Indeed, while the head-and-tail factorization relations, Eqs.~\ref{eq:4nodeheadandtailfactorization}, obey the local and global Markov independence relations entailed by the graphical model, Fig.~\ref{fig:CrossEntropyDecomposition_all}D, leading to the cancellation of all factors associated to non-$ac$-connected subsets in gray in Eq.~\ref{fig:4nodefactorization}, the remaining head-and-tail factors cannot be readily estimated with the empirical distribution $p(.)$.

In particular, the cross-entropy of the
two-collider path
of interest, Fig.~\ref{fig:CrossEntropyDecomposition_all}D, obtained with the head-and-tail factorizations %
corresponds
to\footnote{Indeed, all terms in Eq.~\ref{fig:4nodefactorization} actually cancel two-by-two by construction, {\it whatever their factorized expression}, except for the remaining joint-distribution over all variables, $q(x,z,y,w)\!=\! q(z,y|x,w)\,{{q(x)\,q(w)}}$.}
$H(p,q)\!=\! -\sum p(x,z,y,w) \log q(z,y|x,w)\,{{q(x)\,q(w)}}$.
Then, estimating the $q(.)$ terms with their $p(.)$ counterparts leads to the cross-entropy of a Bayesian graph, Fig.~\ref{fig:CrossEntropyDecomposition_all}E, with a different Markov equivalent class than the ancestral graph of interest, Fig.~\ref{fig:CrossEntropyDecomposition_all}D. A similar discrepancy is obtained with a c-component factorization which leads to the cross-entropy of the Bayesian graph of Fig.~\ref{fig:CrossEntropyDecomposition_all}E without edge $X\to Y$,
corresponding
to a different Markov equivalence class than the previous two graphs, Figs.~\ref{fig:CrossEntropyDecomposition_all}D~\&~E.
These examples illustrate the difficulty to exploit the c-component or head-and-tail factorizations to estimate the likelihood of ancestral graphs including bidirected edge(s).
\section{Node and edge scores based on Normalized Maximum Likelihood criteria}

Search-and-score methods based on likelihood estimates need to properly account for finite sample size, as cross-entropy minimization leads to ever more complex models, resulting in model overfitting for finite datasets.
{\color{black}BIC regularization is valid in the asymptotic limit of very large datasets and leads to the following finite size corrections of the  cross-information terms in the likelihood decomposition Eq.~\ref{LikelihoodDecomposition},
  \begin{equation}
I(\bm{C}) \to I'(\bm{C})= I(\bm{C}) - {1\over 2} \prod_{k=1}^{|\bm{C}|}(1-r_k){\log N \over N}
  \end{equation}
for categorical datasets,  where $r_k$ is the number of categories or levels of the $k$th variable of $\bm{C}$.
  
However, BIC regularization  tends to overestimate finite size corrections, leading to lower recall, in general.}
In order to better take into account finite sample size, we used instead the (universal) Normalized Maximum Likelihood (NML) criterion~\cite[]{shtarkov1987,rissanen2005,kontkanen2007,roos2008}, which amounts to normalizing the likelihood function over all possible datasets with the same number $N$ of samples.

{\color{black}Moreover, as search-and-score structure learning methods need to rely on heuristic rather than exhaustive search, we have implemented a computationally efficient search-and-score method based on the likelihood decomposition of ancestral graphs (Eq.~\ref{LikelihoodDecomposition}) limited to the close surrounding vertices of each node and edge. These node and edge scores, detailed below, extend MIIC's unshielded triple scores to higher-order local information scores including $ac$-connected subsets of vertices with a maximum of two-collider paths.}

{\bf Node score}.~We first used the factorized Normalized Maximum Likelihood (fNML) complexity~\cite[]{kontkanen2007,roos2008} to define a local score for each node $X_i$, which extends the decomposable likelihood of Bayesian graphs given each node's parents, %
Eq.~\ref{BN},
to all non-descendant neighbors, $\mathbf{Pa'}_{\!_{X_i}}$,
\begin{eqnarray}
  {\mathcal{L_{D|G_{\rm X_i}}}}=\;e^{-N.\,{\rm Score}_{\rm n}({X_i})}&=&{e^{-N H(X_i\vert \mathbf{Pa'}_{\!_{X_i}}) }\over \sum_{\vert\mathcal{D'}\vert=N}e^{-N H(X_i\vert \mathbf{Pa'}_{\!_{X_i}})}}\\
  &=&{e^{-N  H(X_i\vert \mathbf{Pa'}_{\!_{X_i}}) -\sum^{q_i}_{j} \log {\mathcal{C}^{r_i}_{n_{j}}}}}\\
&=&{e^{N \sum^{q_i}_{j} \sum^{r_i}_{k} {n_{jk}\over N}\log\left({n_{jk}\over n_{j}}\right) -\sum^{q_i}_{j} \log {\mathcal{C}^{r_i}_{n_{j}}}}}\\
&=&\prod^{q_i}_{j} {\prod^{r_i}_{k} \left({n_{jk}\over n_{j}}\right)^{n_{jk}}\over  {\mathcal{C}^{r_i}_{n_{j}}}}
\label{likelihood_nml1_App}
\end{eqnarray}
where $n_{jk}$ corresponds to the number of data points for which $X_i$ is in its $k$th state and its non-descendant neighbors in their $j$th state, with  $n_{j}=\sum^{r_i}_k n_{jk}$. The universal normalization constant ${\mathcal{C}^{r}_{n}}$ is then computed by summing the numerator over all possible partitions of the $n$ data points into a maximum of $r$ subsets, $\ell_1+\ell_2+\cdots+\ell_r=n$ with $\ell_k\geqslant 0$, %
\begin{align}
{\mathcal{C}^{r}_{n}}&=\sum_{\ell_1+\ell_2+\cdots+\ell_r=n} {{n}!\over{{\ell_1}!{\ell_2}!\cdots{\ell_r}!}}\prod^r_{k=1}\left({\ell_k\over n}\right)^{\ell_k}
\label{Crn_App1}
\end{align}
which can in fact be computed in linear-time using the following recursion \cite[]{kontkanen2007},
\begin{align}
{\mathcal{C}^{r}_{n}}&={\mathcal{C}^{r-1}_{n}} + {n\over{r-2}} {\mathcal{C}^{r-2}_{n}}
\label{Crn_rec_App1}
\end{align}
with %
${\mathcal{C}^{1}_{n}}=1$ for all $n$ and applying Eq.~\ref{Crn_r2_App} below for $r=2$. However, for large $n$ and $r$,  $\mathcal{C}^{r}_{n}$ computation tends to be numerically unstable, which can be circumvented %
by implementing the recursion on parametric complexity ratios  $\mathcal{D}^{r}_{n}=\mathcal{C}^{r}_{n}/\mathcal{C}^{r-1}_{n}$ %
rather than parametric complexities themselves \cite[]{cabeli2020} as,
\begin{eqnarray}
\mathcal{D}^{r}_{n}&=& 1+{n\over{(r-2)\mathcal{D}^{r-1}_{n}}}\label{eq:ComplexityD-NML}\\%\nonumber\\
\log\mathcal{C}^{r}_{n}&=&\sum^r_{k=2}\log\mathcal{D}^{k}_{n}\label{eq:ComplexityC-NML}%
\end{eqnarray}
for $r\geqslant 3$, with ${\mathcal{C}^{1}_{n}}=1$ and $\mathcal{C}^{2}_{n}=\mathcal{D}^{2}_{n}$, which can be computed directly 
with the general formula, Eq.~\ref{Crn_App1}, for $r=2$,
\begin{align}
{\mathcal{C}^{2}_{n}}&=\sum^{n}_{h=0} \binom{n}{h} \left({h\over n}\right)^{h} \left({{n-h}\over n}\right)^{n-h} 
\label{Crn_r2_App}
\end{align}
{\color{black}{or its Szpankowski approximation for large $n$ (needed for $n>1000$ in practice) \cite{szpankowski2001,kontkanen2003,kontkanen2009},
\begin{align}
{\mathcal{C}^{2}_{n}}&=\sqrt{n\pi \over 2} \left(1+{2\over 3}\sqrt{2\over{n\pi}}+{1\over{12n}}+\mathcal{O}\left({1 \over n^{3/2}}\right)\right)\\
&\simeq\sqrt{n\pi \over 2} \exp\left({\sqrt{8\over{9n\pi}}+{{3\pi -16}\over{36n\pi}}}\right)
\label{Crn_approx_App}
\end{align}
}}}

This leads to the following local score for each node $X_i$, which is minimized over alternative combinations of non-descendant neighbors, $\mathbf{Pa'}_{\!_{X_i}}\!\subseteq\mathbf{Pa}_{_{X_i}}\!\!\cup \mathbf{Sp}_{_{X_i}}\!\!\cup \mathbf{Ne}_{_{X_i}}\!$, in the first step of the local search-and-score algorithm (step~1) detailed in the main text, 
\begin{equation}
{\rm Score_{n}}(X_i) = H(X_i|\mathbf{Pa'}_{\!_{X_i}}) + {1\over N}\sum_j^{q_{x_i}}\log\mathcal{C}^{r_{x_i}}_{n_j}\label{eq:nodescore2}
\end{equation}

{\bf Edge scores}.~We then defined several edge scores
to optimize the orientation of each edge, $X\adjacent Y$,
given its close surrounding vertices.

To this end, we first introduced a local score for node pairs which simply sums the node scores, Eq.~\ref{eq:nodescore2}, for each node. The resulting pair scores are listed in Table~2 for unconnected node pairs and for pairs of nodes connected by a directed edge, where $\mathbf{Pa'}_{\!_{X\!\setminus\! Y}}\!=\mathbf{Pa}_{_{X}}\!\cup \mathbf{Sp}_{_{X}}\!\!\!\setminus\! Y$ and $\mathbf{Pa'}_{\!_{Y\!\setminus\! X}}\!=\mathbf{Pa}_{_{Y}}\!\cup \mathbf{Sp}_{_{Y}}\!\!\!\setminus\! X$ with their corresponding combinations of levels, $q_{_{y\!\setminus\! x}}$ and $q_{_{x\!\setminus\! y}}$.

\begin{table}[h]
  \caption{Local scores for node pairs}  
  \hrule
  \smallskip
     \begin{tabular}{lll}
         {\sf\scriptsize Pair score} & {\sf\scriptsize Information} & {\sf\scriptsize fNML Complexity}  \medskip\\
$X\notadjacent Y$ & $H(X|\mathbf{Pa'}_{\!_{X\!\setminus\! Y}}) +H(Y|\mathbf{Pa'}_{\!_{Y\!\setminus\! X}})$ & $~{1\over N}\Big(~~\sum_j^{q_{_{x\!\setminus\! y}}}\log \mathcal{C}_{n_j}^{r_x}+\sum_j^{q_{_{y\!\setminus\! x}}}\log \mathcal{C}_{n_j}^{r_y}~\Big)$ \medskip\\
         $X\to Y$ & $H(X|\mathbf{Pa'}_{\!_{X\!\setminus\! Y}}) +H(Y|\mathbf{Pa'}_{\!_{Y\!\setminus\! X}},X)$ & ${1\over N}\Big(~~\sum_j^{q_{_{x\!\setminus\! y}}}\log \mathcal{C}_{n_j}^{r_x}+\sum_j^{q_{_{y\!\setminus\! x}}r_x}\log \mathcal{C}_{n_j}^{r_y}~\Big)$ \medskip\\
         $X\leftarrow Y$ & $H(X|\mathbf{Pa'}_{\!_{X\!\setminus\! Y}},Y) +H(Y|\mathbf{Pa'}_{\!_{Y\!\setminus\! X}})$ & ${1\over N}\Big(~~\sum_j^{q_{_{x\!\setminus\! y}}r_y}\log \mathcal{C}_{n_j}^{r_x}+\sum_j^{q_{_{y\!\setminus\! x}}}\log \mathcal{C}_{n_j}^{r_y}~\Big)$ \medskip
     \end{tabular}
     \hrule
     \smallskip
    \end{table}

Then, edge scores for directed edges, $X\to Y$ and $Y\to X$, are defined w.r.t.~to the edge removal score, $X\notadjacent Y$, by substracting the pair scores of unconnected pairs to the pair scores of directed edges, leading to the following
edge orientation scores,
\begin{eqnarray}
{\rm Score}(X\to Y) &=& -I(X;Y|\mathbf{Pa'}_{\!_{Y\!\setminus\! X}}) + {1\over N}\Big(~~\sum_j^{q_{_{y\!\setminus\! x}}r_x}\log \mathcal{C}_{n_j}^{r_y}-\sum_j^{q_{_{y\!\setminus\! x}}}\log \mathcal{C}_{n_j}^{r_y}~\Big)\\
{\rm Score}(Y\to X) &=& -I(X;Y|\mathbf{Pa'}_{\!_{X\!\setminus\! Y}}) + {1\over N}\Big(~~\sum_j^{q_{_{x\!\setminus\! y}}r_y}\log \mathcal{C}_{n_j}^{r_x}-\sum_j^{q_{_{x\!\setminus\! y}}}\log \mathcal{C}_{n_j}^{r_x}~\Big)\label{eq:edgescoreasym}
\end{eqnarray}
However, if $r_x\neq r_y$, the fNML complexities of these orientation scores are not identical for Markov equivalent edge orientations between nodes sharing the same parents (or spouses) \cite[]{chickering},  $\mathbf{Pa'}_{\!_{Y\!\setminus\! X}}=\mathbf{Pa'}_{\!_{X\!\setminus\! Y}}=\mathbf{Pa'}$ and $q_{_{y\!\setminus\! x}}=q_{_{x\!\setminus\! y}}$, despite sharing the same conditional mutual information,
\begin{eqnarray}
I(X;Y|\mathbf{Pa'}) &=& {1\over 2}\Big( H(X|\mathbf{Pa'}) +H(Y|\mathbf{Pa'},X) \Big) + {1\over 2}\Big( H(X|\mathbf{Pa'},Y) +H(Y|\mathbf{Pa'})\Big)~~~~~~~~\label{eq:edgeinfosym}
\end{eqnarray}
This suggests to symmetrize the fNML complexities for edge orientation scores by averaging them over each directed orientation, as for the conditional information in Eq.~\ref{eq:edgeinfosym}, leading to the proposed fNML complexity for directed edges given in Table~1 in the main text.

For bidirected edges, the proposed local orientation score accounts for all $ac$-connected subsets in close vicinity of the bidirected edge, which concerns all subsets including either $X$ and any combination (possibly void) of parents or spouses different from $Y$ ({\it i.e.}~corresponding to the information contributions $H(X|\mathbf{Pa'}_{\!_{X\!\setminus\! Y}})$) or  $Y$ and any combination of parents or spouses different from $X$ ({\it i.e.}~corresponding to the information contributions $H(Y|\mathbf{Pa'}_{\!_{Y\!\setminus\! X}})$) or, else, including both nodes $X$ and $Y$ plus any combination of their parents or spouses, corresponding to the following information contribution, $-I(X;Y|\mathbf{Pa'}_{\!_{XY}})$, where  $\mathbf{Pa'}_{\!_{XY}}=\mathbf{Pa'}_{\!_{X\!\setminus\! Y}}\cup \mathbf{Pa'}_{\!_{Y\!\setminus\! X}}$. This last term, $-I(X;Y|\mathbf{Pa'}_{\!_{XY}})$, contains all the remaining information contributions once the bidirected orientation score is given relative to the edge removal score (Table~2) as for the two directed orientation scores, above. Finally, the symmetrized fNML complexity associated with a bidirected edge should be computed with the whole set of conditioning parents or spouses, $\mathbf{Pa'}_{\!_{XY}}$, as indicated in Table~1. Note that this bidirected orientation score becomes also Markov equivalent to the two directed orientation scores, as required, when the nodes share the same parents and spouses, {\it i.e.}~$\mathbf{Pa'}_{\!_{XY}}=\mathbf{Pa'}_{\!_{Y\!\setminus\! X}}=\mathbf{Pa'}_{\!_{X\!\setminus\! Y}}$ and $q_{_{xy}}=q_{_{y\!\setminus\! x}}=q_{_{x\!\setminus\! y}}$ in Table~1.

\section{Toy models}

\begin{figure*}[h!]
\begin{center}
\includegraphics[width=0.7\linewidth]{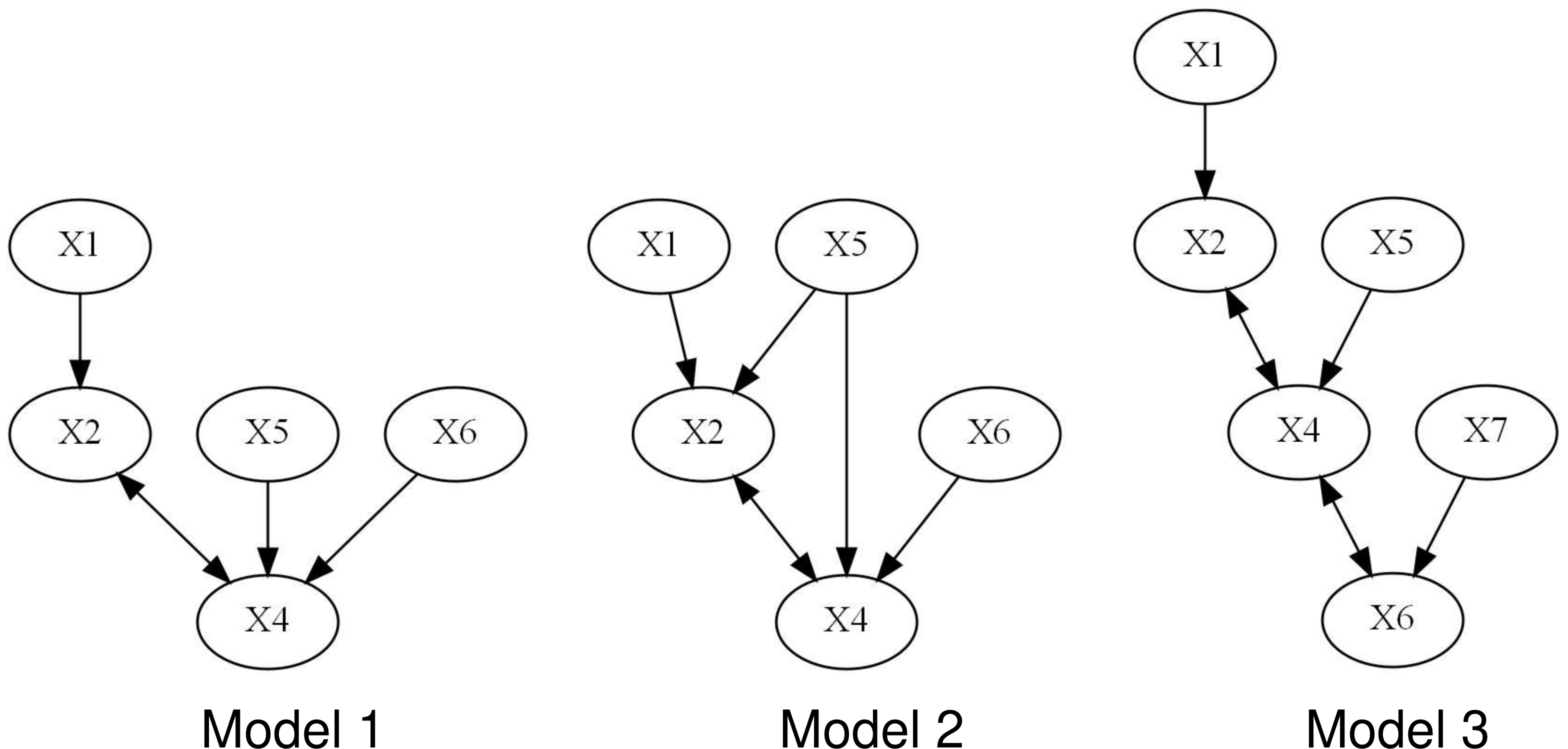}
\end{center}
\caption{\footnotesize{\bfseries Simple ancestral graphs.}}
\label{fig:toy}
\end{figure*}

Fig.~\ref{fig:toy} shows three simple ancestral models used to test MIIC\_search\&score orientation scores (Table~1) to effectively predict bidirected orientations when the end nodes do not share the same parents (Model~1), share some parents (Model~2) or when the bidirected edge is part of a longer than two-collider paths  (Model~3).

The data is generated  from the theoretical DAG using the rmvDAG function in the pcalg package \cite[]{kalisch2012}. Each node follows a normal distribution, and the data is discretized using bnlearn's discretize function using Hartemink's pairwise mutual information method \cite[]{scutari2010}.
For these toy models, the edge orientation scores are computed assuming the correct parents of each node. %
 
The prediction of the edge orientation scores are summarized in Table~3 in \% of replicates displaying directed edges (wrong) or bidirected edge (correct) as a function of increasing dataset size $N$.

\begin{table}[h]
  \caption{\hspace*{0.3cm} Model 1, $X_2\adjacent X_4$ \hspace*{0.3cm} Model 2, $X_2\adjacent X_4$ \hspace*{0.3cm} Model 3, $X_2\adjacent X_4$ \hspace*{0.3cm} Model 3, $X_4\adjacent X_6$}  
  \hrule
  \smallskip
     \begin{tabular}{rrrrrrrrrrrrr}
$N$	& ~~~~~~$\leftarrow$	& $\rightarrow$	& $\leftrightarrow$	& ~~~~~~$\leftarrow$	& $\rightarrow$	& $\leftrightarrow$	& ~~~~~~~~~$\leftarrow$	& $\rightarrow$	& $\leftrightarrow$	& ~~~~~~$\leftarrow$	& $\rightarrow$	& $\leftrightarrow$ \\
1000	& 0	& 100	& 0	& 50	& 42	& 8	& {\color{black}2}	& {\color{black}98}	& {\color{black}0}	& {\color{black}94}	& {\color{black}6}	& {\color{black}0} \\
5000	& 0	& 68	& 32	& 18	& 2	& 80	& {\color{black}22}	& {\color{black}78}	& {\color{black}0}	& {\color{black}62}	& {\color{black}38}	& {\color{black}0} \\
10000	& 0	& 10	& 90	& 0	& 0	& 100	& {\color{black}12}	& {\color{black}70}	& {\color{black}18}	& {\color{black}40}	& {\color{black}16}	& {\color{black}44} \\
20000	& 0	& 0	& 100	& 0	& 0	& 100	& 0	& 0	& 100	& 2	& 0	& 98 \\
35000	& 0	& 0	& 100	& 0	& 0	& 100	& 0	& 0	& 100	& 0	& 0	& 100 \\
50000	& 0	& 0	& 100	& 0	& 0	& 100	& 0	& 0	& 100	& 0	& 0	& 100 \\
     \end{tabular}
     \hrule
     \smallskip
    \end{table}

\end{document}